\documentclass[sn-basic,twocolumn]{sn-jnl}

\usepackage{mathptmx}
\usepackage{graphicx}%
\usepackage{multirow}%
\usepackage{amsmath,amssymb,amsfonts}%
\usepackage{amsthm}%
\usepackage{mathrsfs}%
\usepackage[title]{appendix}%
\usepackage{xcolor}
\usepackage{textcomp}%
\usepackage{manyfoot}%
\usepackage{booktabs}%
\usepackage{algorithm}%
\usepackage{algorithmicx}%
\usepackage{algpseudocode}%
\usepackage{listings}%
\usepackage[font={small,it}]{caption}
\usepackage{subcaption}
\usepackage[T1]{fontenc}
\usepackage{setspace}
\usepackage{tikz}
\usepackage{siunitx} 
\usepackage{comment}

\usepackage{hyperref}
\hypersetup{hyperfootnotes=true}

\graphicspath{{Figs/}}

\begin{document}

\title[Article Title]{In-process \textbf{3D} \textbf{D}eviation \textbf{M}apping and \textbf{D}efect \textbf{M}onitoring (\textbf{3D-DM$^2$}) in High Production-rate Robotic Additive Manufacturing}


\author[1,2]{\fnm{Subash} \sur{Gautam}}
\author*[1]{\fnm{Alejandro} \sur{Vargas-Uscategui}}
\email{alejandro.vargas@csiro.au}

\author[1]{\fnm{Peter} \sur{King}}

\author[1]{\fnm{Hans} \sur{Lohr}}

\author[2]{\fnm{Alireza} \sur{Bab-Hadiashar}}

\author[2]{\fnm{Ivan} \sur{Cole}}

\author*[2]{\fnm{Ehsan} \sur{Asadi}}
\email{ehsan.asadi@rmit.edu.au}

\affil[1]{\orgdiv{Manufacturing}, \orgname{CSIRO}, \city{Clayton}, \postcode{3168}, \state{VIC}, \country{Australia}}

\affil[2]{\orgdiv{School of Engineering}, \orgname{RMIT University}, \orgaddress{\city{Melbourne}, \postcode{3000}, \state{VIC}, \country{Australia}}}

\abstract{\setstretch{1.1}{Additive manufacturing (AM) is an emerging digital manufacturing technology to produce complex and freeform objects through a layer-wise deposition. High deposition rate robotic AM (HDRRAM) processes, such as cold spray additive manufacturing (CSAM), offer significantly increased build speeds by delivering large volumes of material per unit time. However, maintaining shape accuracy remains a critical challenge, particularly due to process instabilities in current open-loop systems. Detecting these deviations as they occur is essential to prevent error propagation, ensure part quality, and minimize post-processing requirements. This study presents a real-time monitoring system to acquire and reconstruct the growing part and directly compares it with a near-net reference model to detect the shape deviation during the manufacturing process. The early identification of shape inconsistencies, followed by segmenting and tracking each deviation region, paves the way for timely intervention and compensation to achieve consistent part quality. }}




\maketitle
\onehalfspacing
\section{Introduction}\label{intro}
Additive manufacturing (AM) is an emerging direct digital manufacturing technology that produces complex free-standing components layer by layer \citep{ngo_additive_2018, iso_isoastm_2021}. The integration of industrial robots with additive manufacturing enhances flexibility, scalability, and the ability to handle complex designs \citep{grau_robots_2021}. Robotic AM (RAM) is growing significantly due to its flexibility and ability to produce complex free-form structures. Metal-based RAM is used ubiquitously to produce a diverse range of near-net metal parts for aerospace, defense, automotive, medical, and construction applications.  Although powder-based fusion AM (PBF-AM) is known for finer precision and resolution to produce zero-tolerance parts, High deposition rate RAM (HDRRAM), such as direct energy deposition AM (DED-AM) and cold spray AM (CSAM), produce near-net objects with a high volume of material per unit time \citep{ralls_solid-state_2023}. 

High Deposition Rate Robotic Additive manufacturing (HDRRAM) processes, such as CSAM, have emerged as promising technologies for large-scale fabrication due to their ability to achieve significantly increased build rates. By enabling the rapid deposition of substantial material volumes, these processes offer clear advantages in terms of throughput and cost-efficiency. However, a critical limitation remains in the form of shape accuracy, which is often compromised by process instabilities inherent to current open-loop control systems \citep{vafadar_advances_2021}. In the absence of real-time feedback, variations in deposition rate and toolpath dynamics can lead to cumulative geometric deviations and surface irregularities. These challenges present a substantial barrier to the broader industrial adoption of HDRRAM, particularly in applications where dimensional precision and repeatability are paramount. Advancing the state of HDRRAM thus necessitates the development of robust, closed-loop control frameworks and in-situ monitoring strategies capable of dynamically identifying and correcting deviations during the build process. 

In such an open-loop system, while single-track studies on flat substrates have been instrumental in guiding the initial selection of process parameters for open-loop HDRRAM systems, they do not adequately capture the complexities inherent in multi-track, multi-layer fabrication. In practical manufacturing environments, each deposited layer serves as the foundation for the next, and any imperfections—such as uneven surface height, residual material buildup, or localized roughness—can propagate and amplify through successive layers. Compounding this issue, inherent process variability and external disturbances make it impractical to estimate and predefine optimal parameters for every layer across an entire build. Therefore, achieving consistent geometric accuracy and build quality demands real-time monitoring and adaptive control strategies that can dynamically respond to evolving deposition conditions.

Even with careful offline tuning of process parameters, HDRRAM processes remain susceptible to geometric deviations such as surface variation, overbuild, and underbuild. These deviations primarily arise from both inherent process fluctuations and limitations in the slicing and toolpath generation procedures. For example, inaccurate estimation of slice height during toolpath planning can result in cumulative height errors. Additionally, the raster-based deposition strategy introduces kinematic challenges, particularly at the ends of scan lines where direction reversals occur. During these transitions, deceleration and acceleration phases increase the nozzle’s residence time, leading to excess material deposition and localized overbuild. To mitigate this, a rounded corner radius is often integrated into the toolpath to smooth transitions and regulate deposition. However, while this strategy may reduce overbuild, an excessively large corner radius can conversely result in under-deposition, compromising dimensional fidelity. Achieving the optimal balance between traverse speed, feed rate, and corner radius is therefore critical to maintaining geometric precision \citep{vargas-uscategui_toolpath_2021}. Yet, these parameters are highly dependent on material properties and machine-specific dynamics, and the resulting geometric distortions introduced in one layer are typically exacerbated in subsequent layers, ultimately impacting the overall accuracy of the final part.

To address these challenges, real-time monitoring during the additive manufacturing process has become increasingly feasible with advancements in vision systems \citep{chen_rapid_2020}. However, effective implementation of real-time monitoring in HDRRAM remains complicated by the minimal standoff distance between the nozzle and the substrate, critical for deposition precision, resulting in a restricted field of view for process sensors. While the system's 6-degree-of-freedom (6DOF) robotic arm enables complex positioning and orientation adjustments of the substrate or nozzle, it is often difficult to position vision sensors close to the deposition area without obstructing the view. This limitation significantly constrains the ability to capture consistent, high-quality surface data, as the limited line of sight and narrow range of viable viewing angles make continuous monitoring challenging. These constraints highlight the need for innovative sensor integration strategies that can accommodate the dynamic nature of HDRRAM and enable comprehensive, real-time feedback for process correction.

Traditionally, defect detection in additive manufacturing has relied on post-process inspection using destructive and non-destructive testing methods, which are both time-consuming and costly \citep{malakizadi_post-processing_2022}. As an alternative, in-process monitoring techniques are being developed to provide real-time, spatiotemporal insights into the manufacturing process by fusing sensor data with toolpath information. By incorporating sensors to capture process signatures—such as temperature, pressure, flow rate, and laser power—during deposition, it becomes possible to detect anomalies as they occur and adjust parameters accordingly \citep{zhang_review_2022}. This real-time insight not only minimizes defect occurrence but also enhances the repeatability and reliability of HDRRAM, representing a critical step toward the development of closed-loop control systems necessary for high-precision, scalable additive manufacturing.

With the advancement of vision systems, real-time monitoring during additive manufacturing processes has become increasingly feasible \citep{chen_rapid_2020}. However, the unique spatial constraints of additive manufacturing, particularly the minimal standoff distance between the nozzle and the substrate, limit the integration of conventional imaging systems. This close proximity, while essential for deposition precision, often obstructs the field of view required for effective monitoring. Even though HDRRAM systems typically utilize 6-degree-of-freedom (6DOF) robotic arms for manipulating the nozzle and substrate, maintaining a clear and stable line of sight remains challenging. These limitations significantly restrict the ability to capture consistent, high-resolution data, thereby impeding accurate real-time surface assessment.

To overcome these challenges, recent efforts have focused on the development of multi-camera, multi-view vision systems that can dynamically accommodate occlusions during the build process. For instance, one study implemented a strategically arranged multi-camera setup designed to maintain visual coverage despite the motion of the build plate during cold spray operations \citep{chew_-process_2024}. This configuration ensured uninterrupted observation of the deposition region from multiple angles, thereby enabling more robust and accurate 3D surface reconstruction. By fusing information from different perspectives, such systems significantly enhance the reliability of temporal tracking and surface characterization, which are crucial for effective quality assurance in dynamic and geometrically complex additive manufacturing environments.

The effectiveness of such real-time monitoring strategies depends significantly on the choice of sensing modality, which is dictated by the specific characteristics of the build to be measured. Active vision sensors, particularly 2D laser profilers, have demonstrated superior speed and accuracy over 3D cameras in capturing fine geometric features such as layer height and width, wall angles, and surface roughness. These profilers generate high-speed, accurate 2D scans of the surface, producing point clouds that reflect the evolving topography as the sensor moves relative to the part. This makes them especially suitable for high-rate deposition processes, where rapid feedback is required. Repeating these scans during the build produces a time series of surface profiles that can be analyzed to detect deviations or trends in deposition quality.

However, integrating these point clouds into a coherent representation of the build geometry presents its own challenges. The continuously evolving nature of the AM surface means that no two scans are identical, requiring efficient strategies for storing, processing, and fusing unstructured 3D data. Historically, this has been hindered by the significant memory demands of volumetric data storage. Recent advancements in computational power, memory efficiency, and sparse volumetric grid structures have significantly improved the performance of voxel-based representations. These developments have enabled more effective volumetric fusion techniques, originally used with RGB-D sensors, to be adapted for AM applications. Nevertheless, most existing fusion approaches remain tailored to specific sensor configurations, underscoring the need for generalized, sensor-agnostic frameworks that can support diverse data modalities while meeting the real-time constraints of HDRRAM systems.

The study aims to address the critical challenges of in-process geometrical monitoring in RAM. First, we tackle the issue of measuring a full 360$^{\circ}$ view of deposition amidst multi-directional nozzle movements and field-of-view obstructions by proposing a vision system with three laser profilers scanning the deposition area in a triangle arrangement. Spatio-temporal point clouds from various cameras are registered in real-time into a global frame using known homogeneous transformations obtained through hand-eye calibration \citep{gautam_streamlined_2025} and the robot's forward kinematic solution. Second, we address the complexity of capturing and updating an accurate 3D shape of the print in real-time; we fuse all measurements on the part surface by incorporating volumetric fusion with adaptive weights for stationary and growing parts of the dynamic print.
The next challenge arises from the near-net-shaped characteristic of HDRRAM, which is why we cannot use a CAD model as a reference for deviation mapping; certain deviations from the intended geometric dimensions specified in the CAD model are permissible.  To treat this, we propose a novel near-net-shape reference model created via Gaussian approximation modeling of deposition. This model serves as an accurate reference for generating 3D deviation maps in-process by comparing it with the one obtained by real-time 3D reconstruction of deposited surfaces. Finally, the deviation map is automatically analyzed to detect overbuilds and underbuilds as the primary geometric defects. The method also tracks growth rates or self-compensation locally and globally, enabling the tracking of geometric defects within and between layers. 

The following points are the key contributions of this paper.
\begin{itemize}
\item {\it Multi-sensor vision system}: 3D vision system using multiple 2D-laser profilers for real-time scanning of the high-rate AM, eliminating field of view obstruction due to multidirectional tool movement.
\item {\it Volumetric fusion with adaptive weight}: Truncated signed distance field volumetric fusion with adaptive weight strategy for dynamic scene update (growing part) by tracking active deposition areas.
\item {\it Geometric defect detection and segmentation}: Automated geometric defect detection and segmentation by comparison with near-net reference model based on signed distance comparison.
\item {\it Defect tracking}: Track segmented defects over multiple layers and identify self-compensation or amplifications.
\item {\it Experimental study}: Experimental study and validation with multi-track multi-layer complex geometry.
\end{itemize}

The remainder of the paper is organized as follows: \autoref{sec: related} reviews the relevant literature to establish the theoretical framework and highlight existing research gaps. \autoref{sec: 3D-DM2} outlines the overall framework and design of 3D deviation mapping and defect monitoring, discussing four key pillars of the method presented in consecutive sections 4, 5, 6 and 7. \autoref{sec: RCSAM} presents the current general AM process, such as robotic cold spray AM. \autoref{sec: pcd acquisition} explains more about the 3D vision system and hand-eye calibration, followed by \autoref{sec: dynamic}, which discusses about incremental surface reconstruction method. The deviation mapping and monitoring method is explained in \autoref{sec: deviation}. \autoref{sec: experiment} illustrated the experimental results and analysis of the proposed methods on 3D deviation mapping and defect monitoring system. In-depth discussions about the methodology and practical implications are presented in \autoref{sec: discussion}. Finally, \autoref{sec: conclusion} concludes the paper, summarizing the key contributions and suggesting directions for future research.

\begin{figure*}
    \centering
		\includegraphics[width=0.9\linewidth]{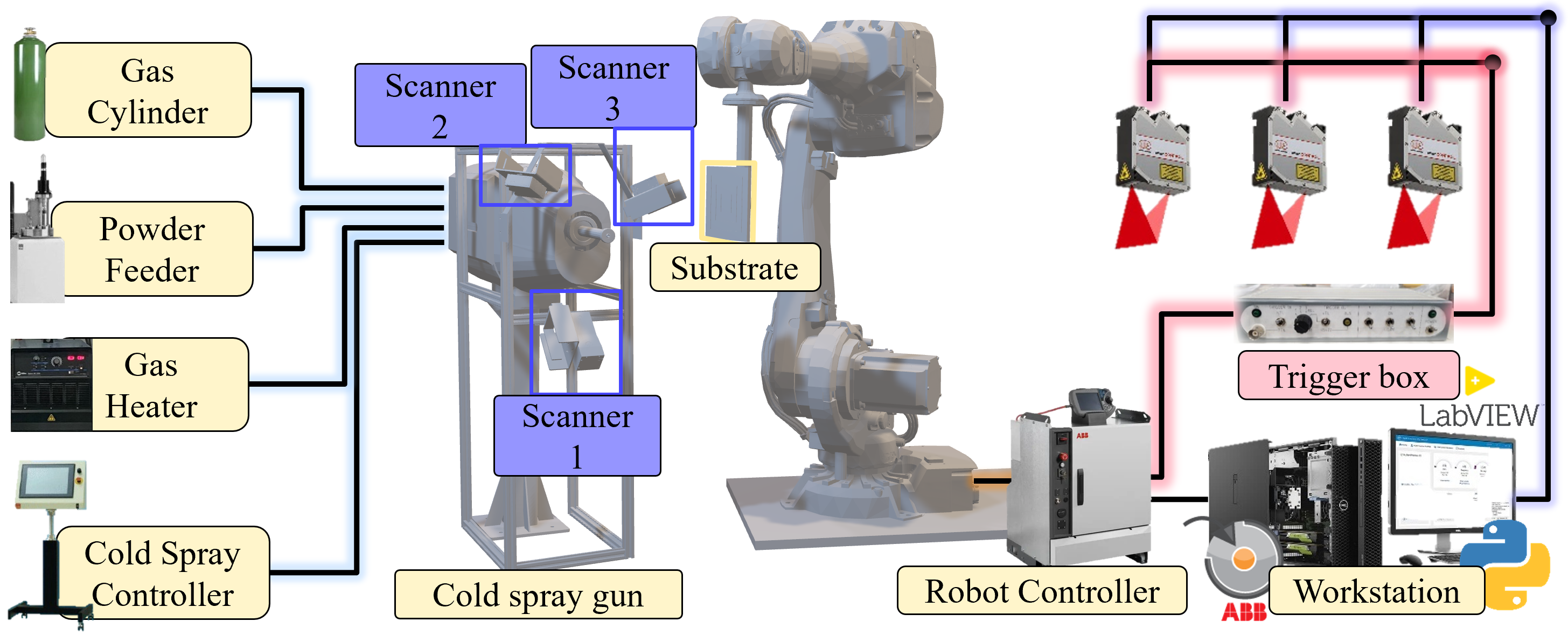}\vspace{0.2cm}
    \caption{Robotics Cold Spray System Equipped with a novel 3D vision system and method using multiple 2D lasers.}
    \vspace{-5mm}
    \label{fig: setup}
\end{figure*}

\section{Related Works}\label{sec: related}
While significant progress has been made in monitoring additive manufacturing processes, much of the existing research has focused on detecting internal material defects such as pores and cracks. However, surface defects i.e. irregularities in geometry, layer height inconsistencies, and surface roughness, can equally compromise the structural integrity and functionality of the final part. Addressing these external defects in real time remains a critical yet underexplored area. The following section reviews related work, underscoring the gap in surface defect detection and the need for improved geometric monitoring techniques.

Defects can be divided into surface defects visible on the outer surface and internal defects not visible, i.e., pores, cracks, and voids \citep{abouelnour_-situ_2022, grasso_process_2017}. Internal defects can be attributed to mechanical issues, while geometric defects cause shape deviation and affect surface quality \citep{abouelnour_-situ_2022}. Hence, geometric accuracy is as essential for parts to be commercially viable as for mechanical properties. Active or passive vision-based monitoring systems can detect surface defects using geometric process signatures \citep{mbodj_review_2023}. 
Contextual information extraction is easier with a machine vision system. It is followed by a series of image processing steps to detect the defects. Although machine vision-based systems are cost-effective, environmental factors, i.e., illumination and light conditions, affect the detection quality. In contrast, \citep{lin_online_2019} proposes laser-based monitoring with active laser light, which directly obtains precise distance information of surface topography as a point cloud.

The surface topography measurement based on laser scanning provides high-resolution data and high-frequency acquisition. The laser scanner directly outputs the distance from the sensor to the object, providing a 3D point cloud of surface \citep{lafirenza_layerwise_2023}. 
\citep{chakraborty_optimal_2017} demonstrated the importance of in-process monitoring and control for optimal control of build height in a single-track CSAM experiment. \citep{heralic_height_2012} measured the quality of the surface in WAAM using a laser scanner and controlled the height of the bead by an iterative learning control. \citep{lin_online_2019, lyu_online_2021, li_geometrical_2021} used a laser profiler to measure the surface after depositing each layer in an extrusion-based Fused Filament Fabrication (FFF). \citep{tang_situ_2019} used a laser profiler to measure the surface topography and flatten each layer by applying a milling and filling strategy for WAAM. \citep{garmendia_development_2019} used a structured light scanner to measure surface layer-by-layer to control part height error and local height deviations in laser metal deposition (LMD) AM. \citep{huang_rapid_2022} developed an inspection system based on a laser profilometer and converted a point cloud to a topographic image to classify defects into different categories, that is, normal, bulgy, dented, and pores. \citep{lin_online_2019} proposed a surface quality monitoring method in extrusion-based AM to detect overfilling and underfilling of extrusion material every few layers after depth image conversion. \citep{lyu_online_2021} developed a process monitoring and control system for extrusion-based Fused Filament Fabrication AM (FFF-AM) to detect and compensate for under-extrusion and over-extrusion of material from the rasterized image in each layer. Most methods use laser profilers to collect 3D point cloud data, although rasterization is used to transform them into ordered depth images, reducing dimensionality \citep{lyu_online_2021, huang_rapid_2022}. The methods generally work by stopping the deposition process to obtain the surface topology of the whole deposition. The disruption in the deposition process increases the printing time. It is also impractical for continuous and complex deposition processes with non-strict layer-by-layer deposition, as in CSAM. Continuous data acquisition and processing require real-time monitoring. Also, the point cloud should be updated incrementally to reconstruct the latest surface topology. In such a scenario, scanning each track during deposition is preferred to cater for the continuous toolpath without sensor intermittence. The above methods relied mainly on layer-wise measurement and did not provide real-time monitoring that can capture inter-intra-layer data. Experimental studies often involved complex shapes; however, single-track and thin-walled structures were studied \citep{ye_predictions_2023, ikeuchi_neural_2019, ikeuchi_data-efficient_2021}. In addition, most methods collected data after the deposited layer by stopping the deposition; however, it increases the production time and is not feasible for processes that cannot be stopped, such as CSAM \citep{chew_-process_2024}. 
Collectively, these studies demonstrate the importance of geometric defect monitoring in additive manufacturing, yet surface-level inspection remains an open challenge,especially in real-time settings. The following section introduces our method, which specifically targets this gap through a novel integration of spatial–temporal data and multi-view geometry using 3D deviation mapping and a defect monitoring framework.


\section{3D Mapping and Monitoring in AM}\label{sec: 3D-DM2}
\begin{figure*}[t]
	\centering
		\includegraphics[width=0.9\linewidth]{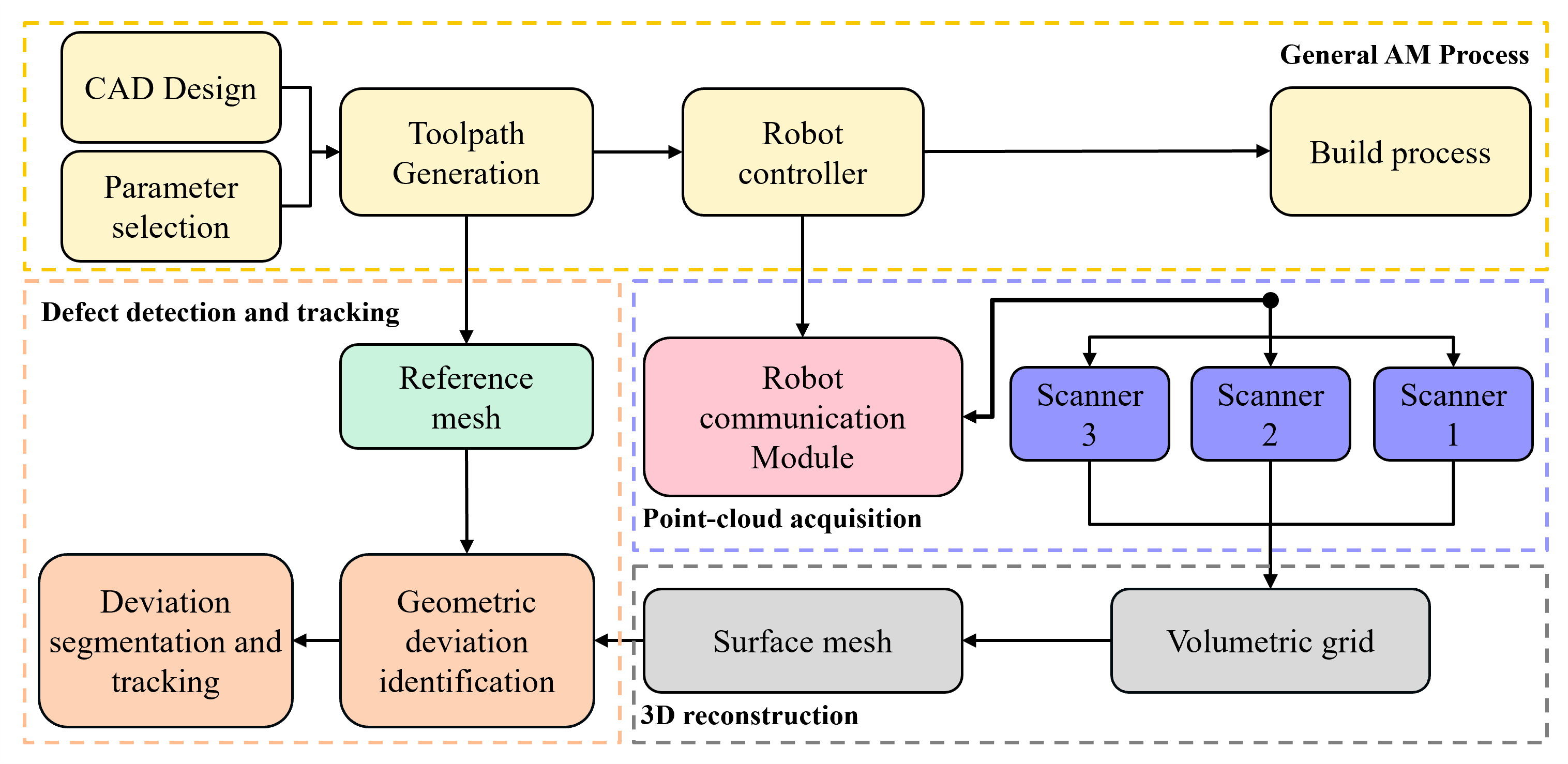}
	\caption{3D-DM$^2$: overall design of 3D vision system, 3D shape reconstruction, 3D deviation mapping and defects monitoring.}
	\label{fig: Flowchart}
\end{figure*}

\autoref{fig: Flowchart} illustrates the proposed 3D Deviation Mapping and Defect Monitoring framework, 3D-DM$^2$, and its four key pillars. The upper box shows the general offline AM process of tool-path generation from the given CAD model, which the robotics cold spray will execute, and is represented in \autoref{sec: RCSAM}. The other three modules on the bottom are executed online in real-time alongside the cold-spray part printing. They are explained in \autoref{sec: pcd acquisition}, \autoref{sec: dynamic}, and \autoref{sec: deviation}. 

\textbf{Robotic cold spray AM: }
The general AM process is initiated by designing a toolpath based on a sliced CAD model and toolpath parameters. Then the AM process is executed by depositing material in a layer-by-layer approach to produce a part. Further details about the existing open-loop system are provided in \autoref{sec: RCSAM}.

\begin{figure}
    \centering
    \begin{subfigure}[b]{0.33\columnwidth}
         \centering 
         \includegraphics[width=\columnwidth]{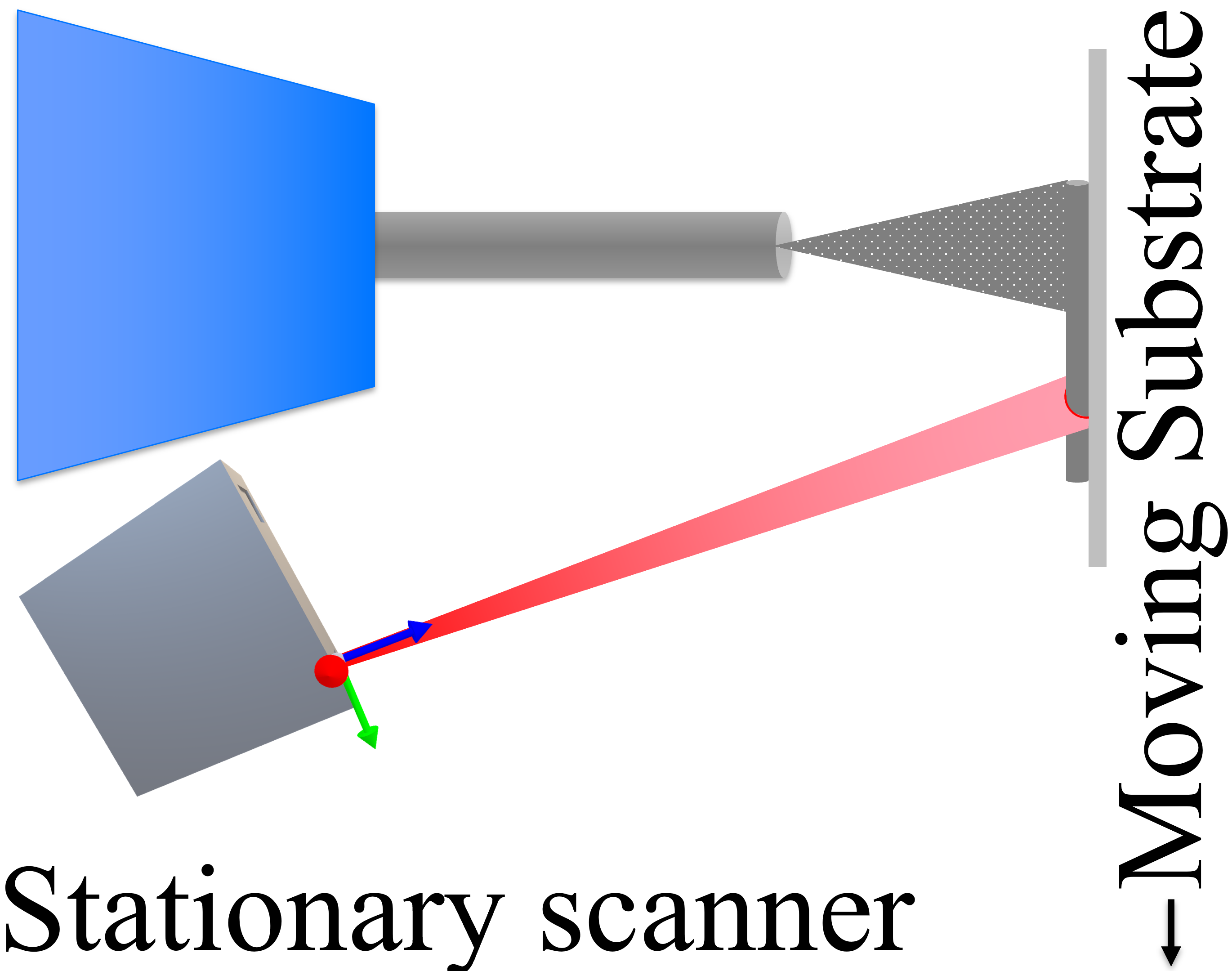}
         \caption{}
         \label{fig: moving_substrate}
    \end{subfigure}
    \begin{subfigure}[b]{0.32\columnwidth}
         \centering 
         \includegraphics[width=\columnwidth]{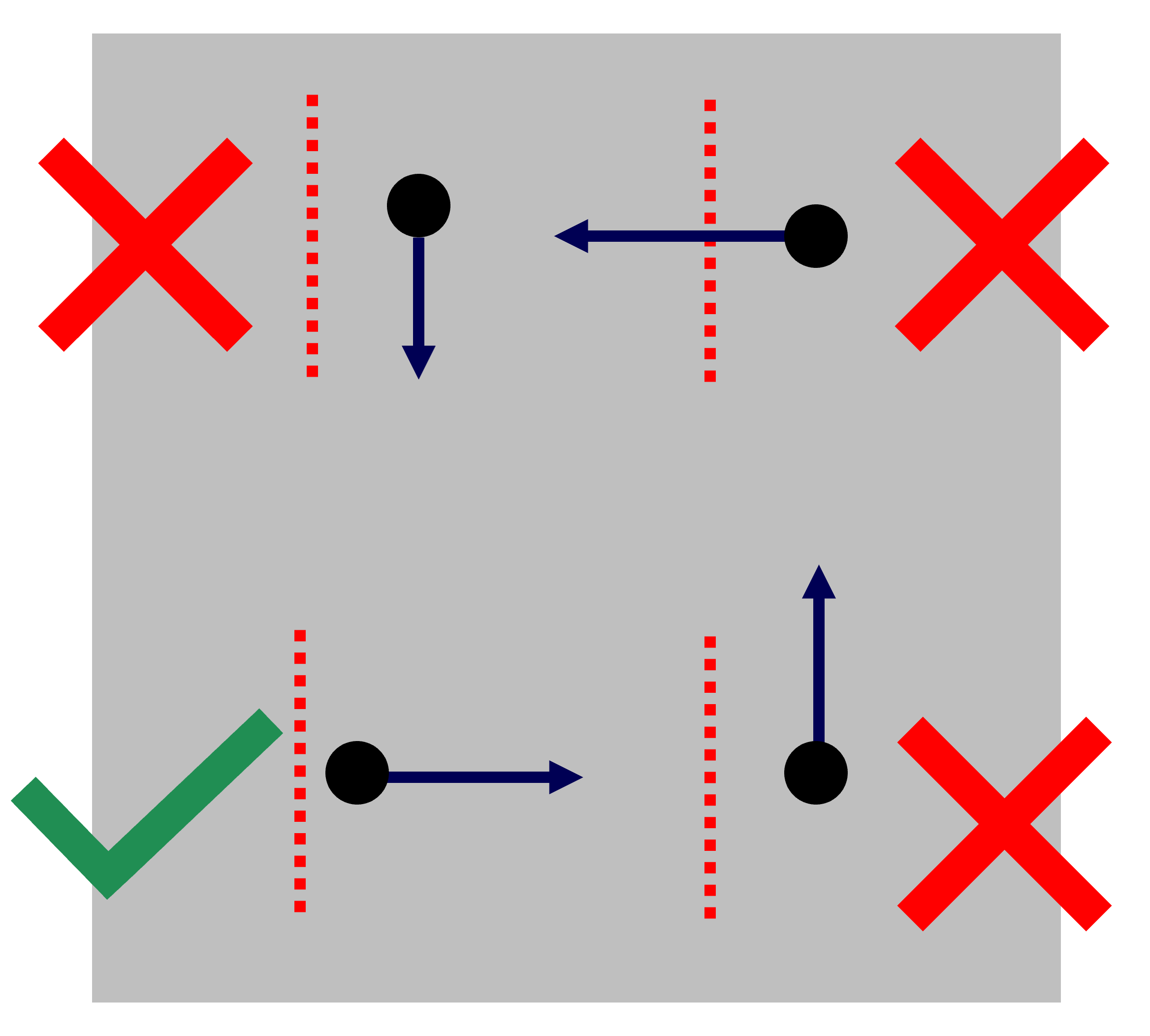}
         \caption{}
         \label{fig: no_multi_direction}
    \end{subfigure}
    \begin{subfigure}[b]{0.33\columnwidth}
         \centering 
         \includegraphics[width=\columnwidth]{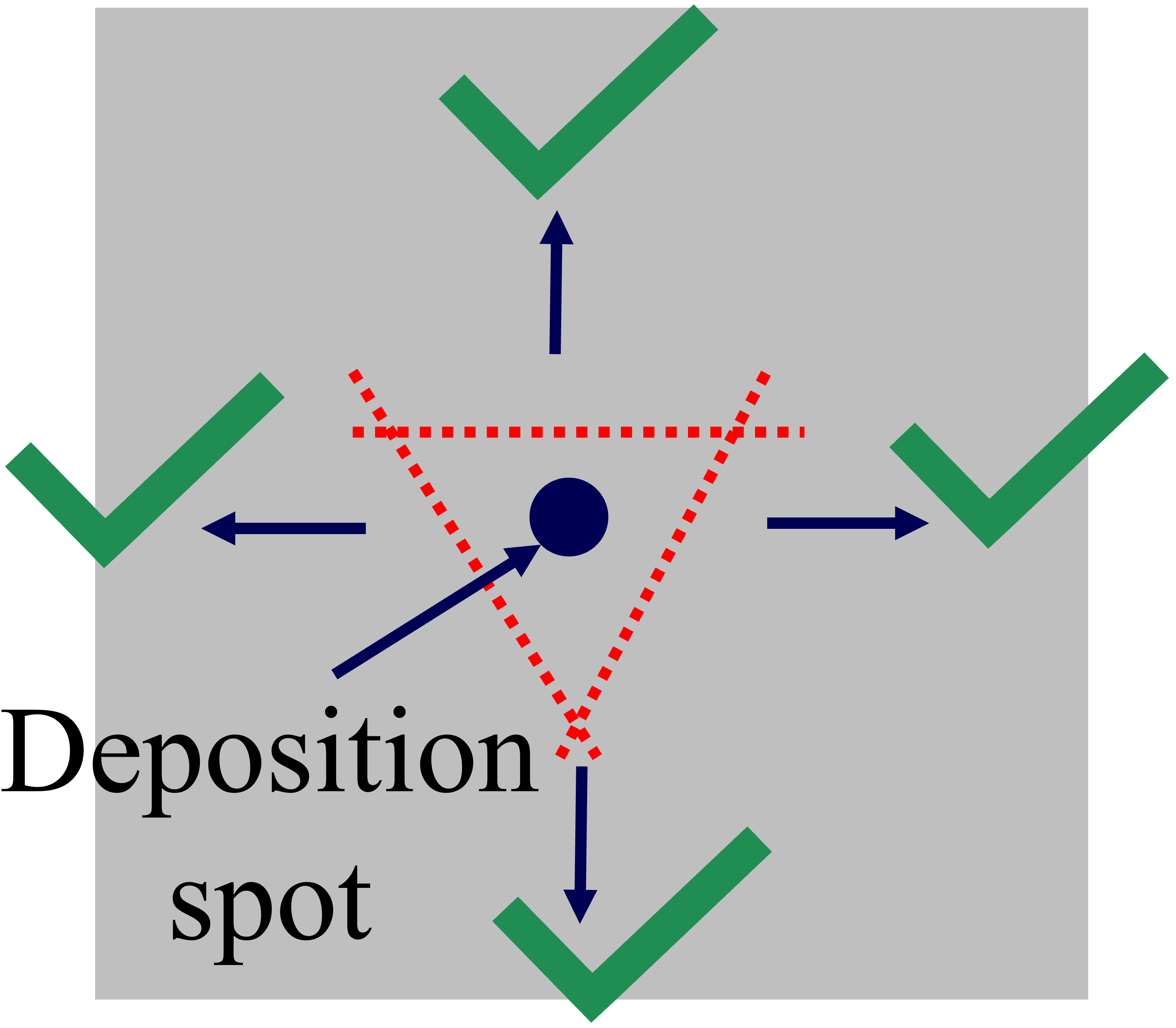}
         \caption{}
         \label{fig: multi_direction}
    \end{subfigure}
    \caption{Multi-sensor vision system for eliminating FOV obstruction due to tool multidirectional movement in AM. (a) A stationary scanner acquires the profile of a moving substrate. (b) A single profile sensor is not able to capture multidirectional deposition. (c) Multiple arrangements of scanners can capture multi-directional deposition. }
    \label{fig: multi_sensor_}
\end{figure}

\textbf{3D vision fusing multi-2D-profilers: }
To overcome the challenge of capturing a full 360{$^o$} views of local depositions in real-time during substrate movement, we propose a sensory system with multiple 2D laser scanners (at least three, forming a triangle) in different directions, as shown in \autoref{fig: multi_sensor_}, to monitor the complete build, \autoref{sec: pcd acquisition}.

\textbf{3D dynamic surface reconstruction: }
This method overcomes the occlusion effects caused by the line-of-sight effect caused by using one sensor to observe from a fixed angle. We propose fusing all measurements from the part surface by incorporating volumetric fusion with adaptive weights for stationary and growing parts of the dynamic print. This method simplifies the challenge of capturing print surfaces in real-time. \autoref{sec: dynamic} elaborates more on this topic.

\textbf{3D deviation mapping, defects detection, and tracking: }
To account for permissible geometric tolerances in near-net-shape HDRRAM, we propose a novel near-net-shape reference model by adapting Gaussian approximation modeling of deposition over the toolpath. This model is converted to a 3D mesh, serving as an accurate reference for generating a 3D deviation map by comparing it with the one obtained by a real-time 3D vision system and surface reconstruction. The deviation map is automatically analyzed to detect geometric defects and to track growth rates or self-compensation locally and globally. This topic is explored in greater depth in \autoref{sec: deviation}.
 
 

 \begin{figure}[!t]
	\centering
		\includegraphics[width=0.99\columnwidth]{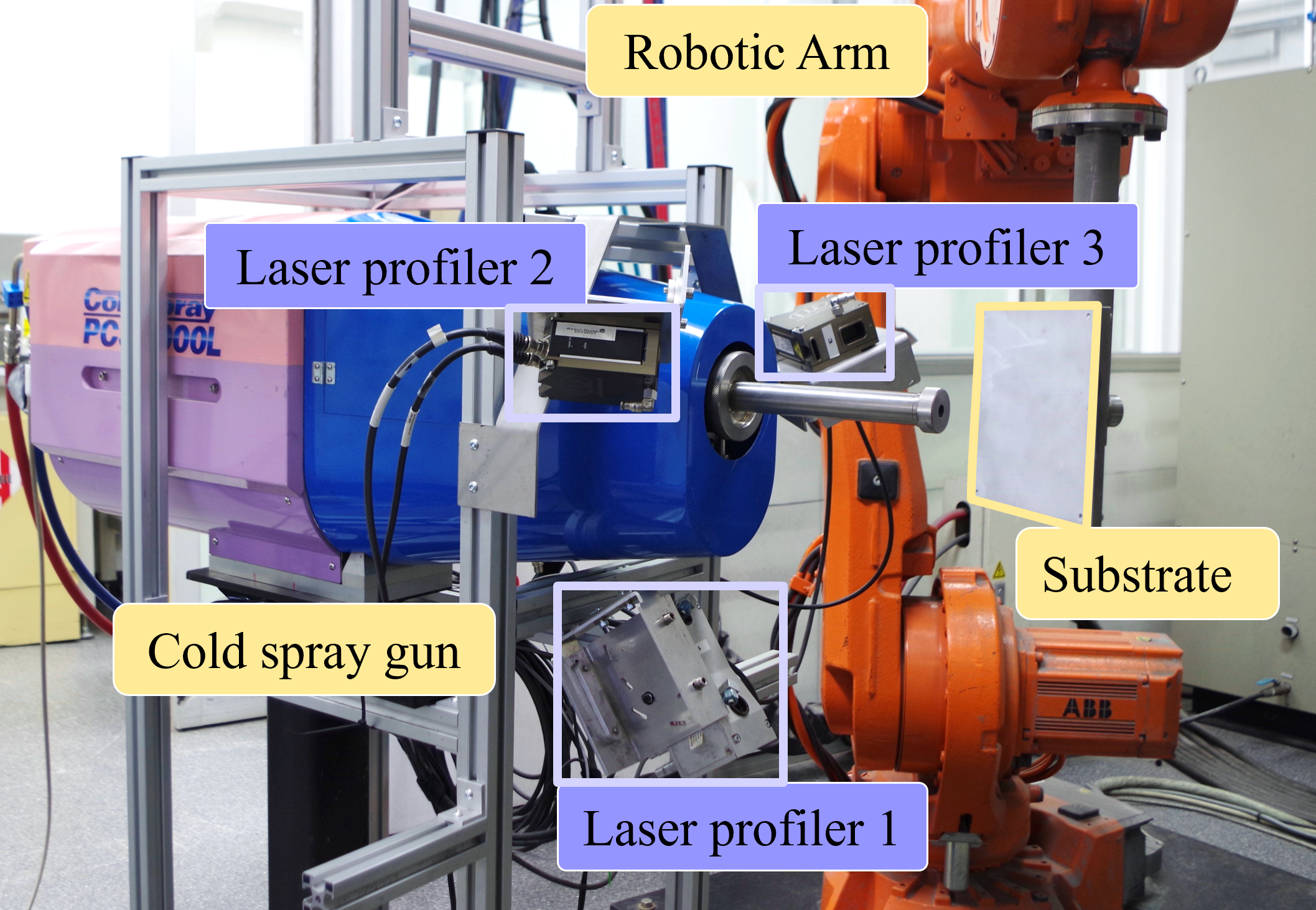}
	\caption{Experimental robotics CSAM setup, multi 2D profilers sensory system and coordinate frames}
	\label{fig: coordinate}
\end{figure}


\section{Robotic Cold Spray additive manufacturing}\label{sec: RCSAM}
The study was conducted in a cold spray lab at the Commonwealth Scientific and Industrial Research Organization (CSIRO), Clayton. The system consists of the cold spray gun, heater, powder feeder, industrial robot, and gas supply, \autoref{fig: setup}. In the current configuration, the cold spray gun is stationary, whereas the robot moves the substrate on a programmed path, known as a toolpath. The cold spray is a plasma Giken PCS-1000L comprising a heater and a gun integrated into a single unit. The maximum chamber temperature is 1000$^o$C, and the maximum gas pressure is 5 MPa.  
The cold spray system passes the gas from the nozzle, which is heated to achieve supersonic speed. After maintaining a stable flow, the powder material is deposited at a high rate on the moving substrate to print a part.

To generate a toolpath for 3D printing, the CAD model is sliced into multiple layers using specific slicing software. Generic 3D printing utilizes material flow control to stop and start deposition, creating a skip movement in the toolpath. However, for continuous material flow, such a toolpath is not suitable. The continuous toolpath for this case study was generated by Continuous3D software (C3D) (developed in-house, CSIRO) (patent no WO2021016666) \citep{king_method_2021}.

After generating the toolpath, the robot program is loaded into the controller. For the CSAM process, it is essential to maintain a constant gas flow during the system warm-up procedure. Once the warm-up procedure is finished, the powder feeder is turned on to allow the powder particles to enter the gas stream. The robot moves the substrate/nozzle to deposit material in layers until the desired 3D part is built. In the current open-loop system, the operator constantly observes the overall part growth. This is replaced by a vision system that captures the part growth constantly, which is discussed in the next section.

\section{3D Vision System and Methods via Fusing Multi 2D Profilers}\label{sec: pcd acquisition}
In the current open-loop cold spray AM, the vision system was integrated to capture part growth. However, a single laser scanner can not capture the complete 360$^o$ view of the build due to obstruction from the cold spray nozzle. 
As shown in \autoref{fig: coordinate}, a multi-directional toolpath is often necessary to build the 3D object. In the proposed system, scans of the local deposition are acquired on the fly by multiple 2D laser profilers placed around the spray nozzle or extrusion system, i.e., a cold spray gun focusing on the build plate. Each laser profiler works according to the optical triangulation principle, also known as a laser light triangulator.  
Three micro epsilon ScanCONTROL laser profilers were used to create a multi-scanner arrangement. Laser profilers are connected to PCs via a Gigabit Ethernet connection and connected to the robot controller via an external trigger box to receive digital input for triggering the profiler. 

The robot controller triggers data acquisition and synchronizes data acquisition using an interrupt subroutine. When the external trigger box receives a trigger signal, it cascades the signal to each profiler at 20 ms intervals to avoid interference from each other. Although the scanner allows high-frequency triggering, the frequency is limited to 10 Hz due to the cascading hardware in the trigger box. The robot controller is connected to the PC via a LAN interface to communicate with the RobotStudio program. The program transmits application data between computers using TCP/IP or UDP protocol.



\subsection{Multi-sensor hand-eye calibration}

Each scanner acquired measurements in its specific coordinate frame ($\{L_1\}$, $\{L_2\}$, $\{L_3\}$). Then, with the help of transformations obtained by hand-eye calibration, these measurements are projected and merged into the coordinate frame of the work object, $\{O\}$, placed in the bottom left corner of the plate. 

A homogeneous transformation correlating any observed point in a 2D profiler to its corresponding point on the calibration plane is expressed as follows:
\begin{equation}    
^OP^j=\hspace{1mm}^O_{L_j}{\textit{\textbf{T}}}\hspace{1mm}^{L_j}P^j =\hspace{1mm}^O_B{\textit{\textbf{T}}}\hspace{1mm}^B_{L_j}{\textit{\textbf{T}}}\hspace{1mm}^{L_j}P^j.
\label{eqn: transformation}
\end{equation}

Here, ${^{L_j}}P^j$ encompasses all points acquired by a scanner $j$, and ${^O}P^j$ represents the projection of the entire laser beam onto the workobject plane. The kinematic chain involves the transformation from the robot base frame to the work object, specifically the substrate plate, denoted $^O_B{\textit{\textbf{T}}}$, determined by the forward kinematics of the manipulator in pose $i$, and the transformation from the sensor $j$ to the robot base frame $^B_{L_j}{\textit{\textbf{T}}}$ which is unknown. Thus $^B_{L_j}\textit{\textbf{T}} \in \mathbb{R}^{4\times4}{}$ that describes the location of sensor coordinate ${\textbf\{L_j\}}$ to robot base ${\textbf\{B\}}$ is calculated via a hand-eye calibration method developed in-house \citep{gautam_streamlined_2025}.

\section{Dynamic Surface Reconstruction}\label{sec: dynamic}
A unified spatio-temporal point cloud is obtained by collecting and projecting surface profile measurements into the work object coordinate frame. We propose fusing all 3D point measurements from the part to generate a smooth surface by incorporating volumetric fusion with adaptive weights for stationary (inactive region) and growing parts (active region) of the dynamic print. This addresses the complexity of capturing dynamic surfaces in real-time. The unordered points are voxelized using the VDB (Volumetric Dynamic B+tree) data structure for faster traversal. The method constantly updates the volumetric grid as points are acquired based on tracking the active deposition area. Surfesh mesh can be extracted at any point from the volumetric grid using the marching cubes method.

Volumetric fusion involves integrating sensor data into a voxel-based representation, such as a Truncated Signed Distance Function (TSDF), and its computational cost is influenced by factors like sensor data density, voxel resolution, and update frequency. The most intensive operation is ray casting to update affected voxels, which can be demanding in high-resolution reconstructions. However, the use of efficient libraries like OpenVDB, which supports sparse voxel storage and multi-threaded processing, significantly reduces the memory and compute overhead by focusing computation only on active regions of the grid.

To achieve real-time monitoring at 10 Hz (one frame every 100 ms), the system is designed to process each incoming frame within the available time budget. This is feasible because the laser profiler produces 640 points per frame per scanner, keeping the per-frame computational load low. Multi-threaded ray casting and OpenVDB's sparse updates ensure that only relevant voxels are updated, allowing the system to consistently meet real-time requirements, even with a 100 mm³ voxel grid.

Latency in this workflow can emerge from TSDF integration, data preprocessing, and memory access overhead. While current performance supports real-time operation, increasing the point density, voxel resolution, or update acquisition rate may introduce delays. The system’s speed scales sub-linearly with such increases, meaning higher spatial or temporal fidelity can raise computational demands disproportionately. For instance, halving voxel size increases voxel count by a factor of eight, and doubling acquisition rate halves the processing time per scan. These factors highlight the trade-off between spatial and temporal resolution, reinforcing the need for adaptive strategies like dynamic resolution, region-of-interest updates, or GPU acceleration to sustain real-time monitoring under changing operational conditions.

\begin{figure}
    \centering
    \begin{subfigure}[b]{0.48\columnwidth}
         \centering 
         \includegraphics[width=0.99\columnwidth]{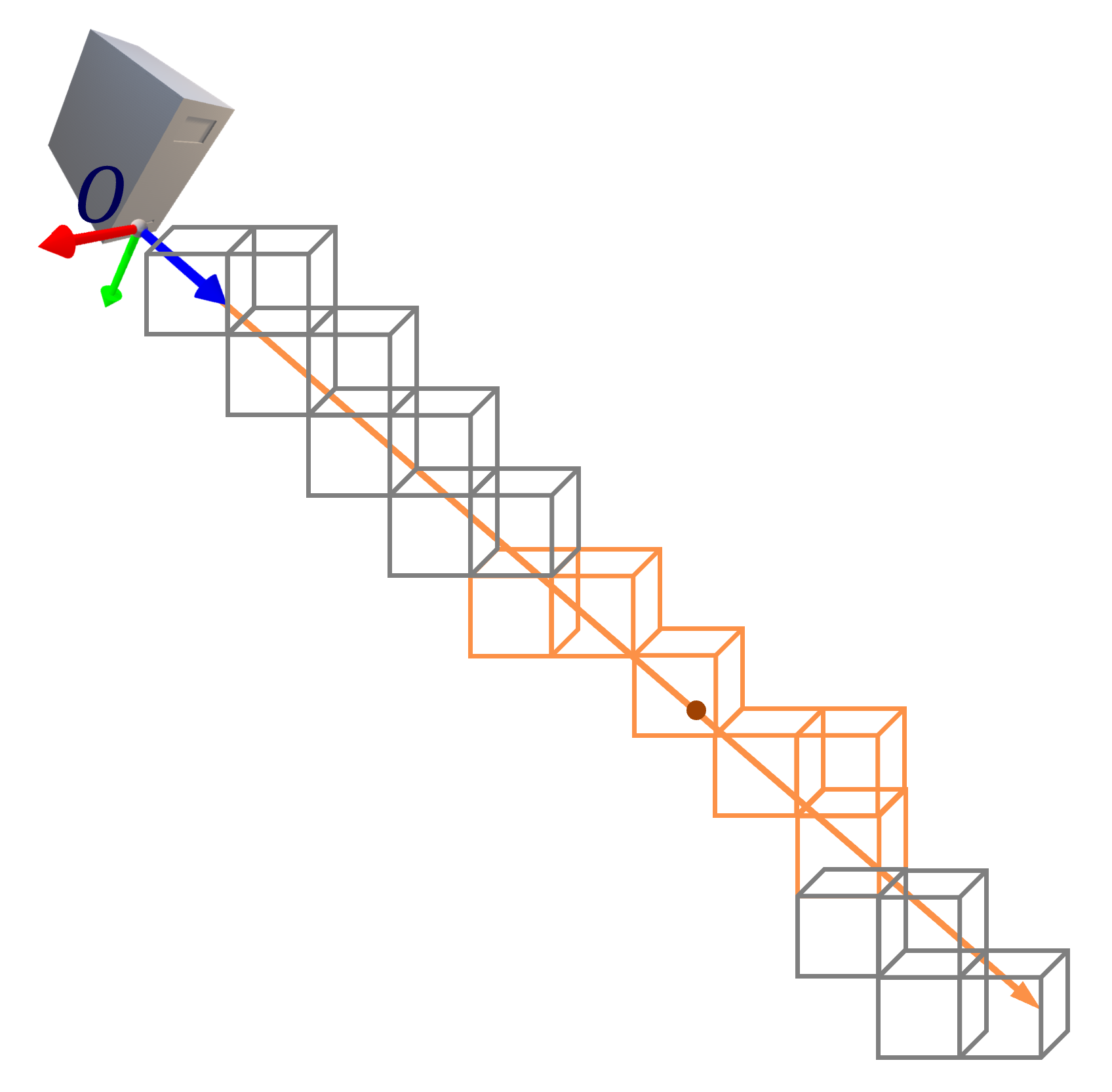}
         \caption{}
         \label{fig: ray_casting}
    \end{subfigure}
    \begin{subfigure}[b]{0.48\columnwidth}
         \centering 
         \includegraphics[width=0.99\columnwidth]{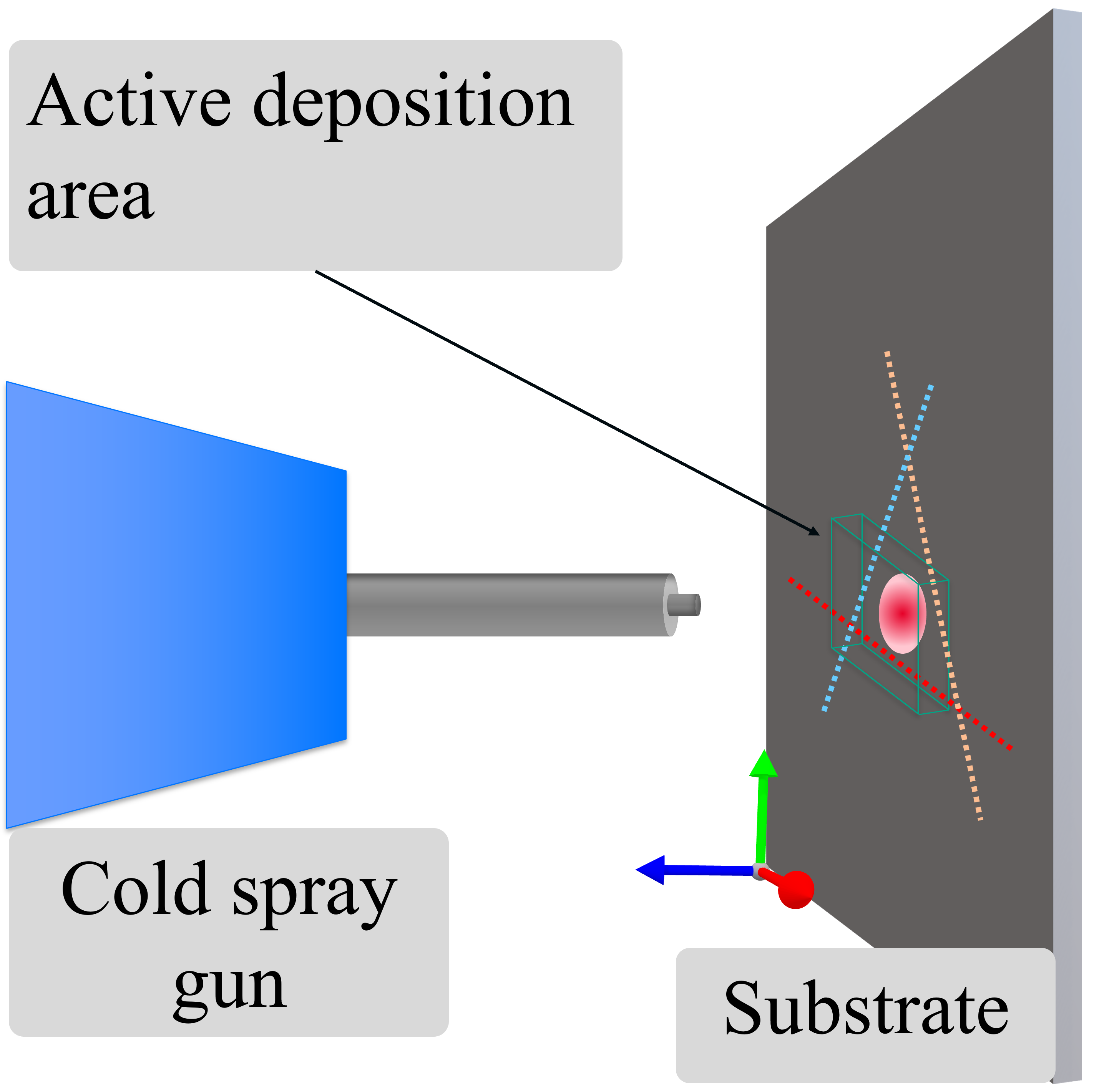}
         \caption{}
         \label{fig: deposition area}
    \end{subfigure}
    \caption{ (a) selected voxels (orange) truncated by a distance based on ray casting, (b) Action area of deposition where the material is being added for dynamic scene update, and Inactive area of deposition used for smoothing noise in the reconstructed surface.}
    \label{fig: ray_casting and deposition area}
\end{figure}

\begin{figure*}[t]
	\centering
		\includegraphics[width=0.99\linewidth]{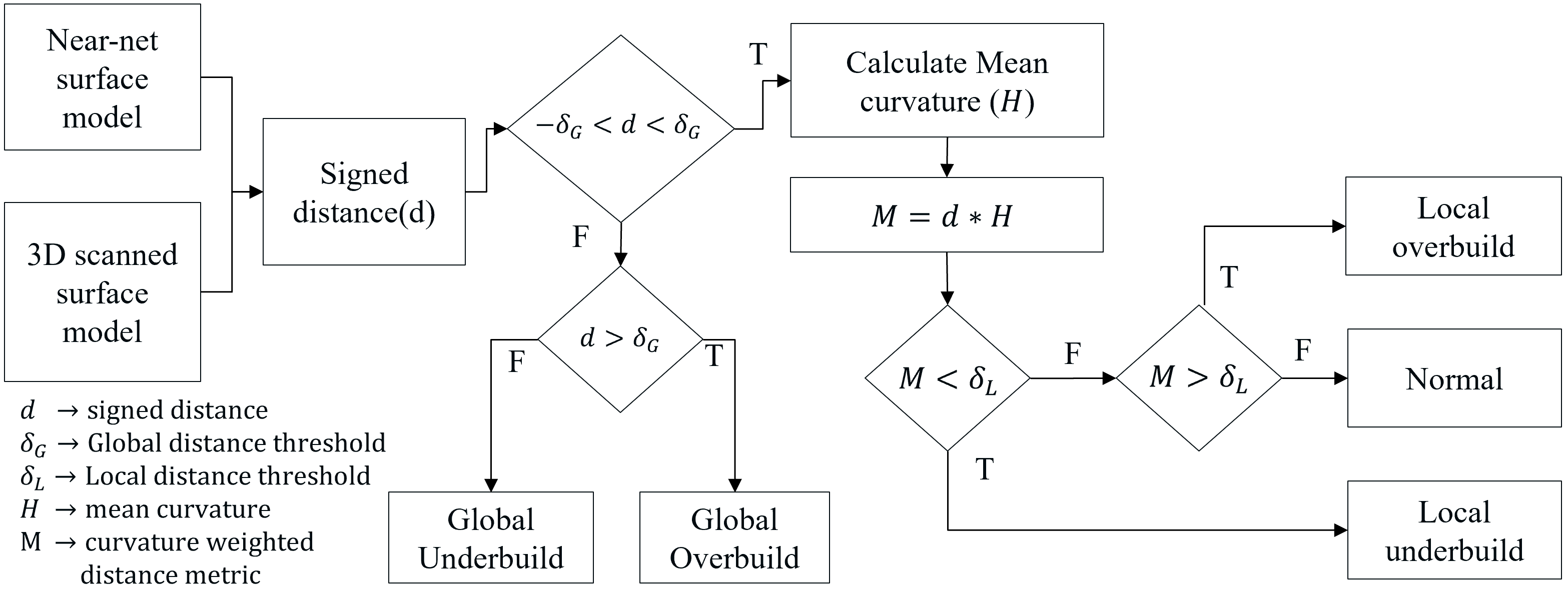}
    \vspace{2mm}
	\caption{3D-DM$^2$: An online defect detection algorithm to determine local deviation or global deviation (part height deviation) in AM.}
	\label{fig: defect detection}
\end{figure*}

\subsection{Volumetric fusion with adaptive weights}\label{fusion}

\citep{curless_volumetric_1996} initially introduced the truncated signed distance field (TSDF) for volumetric 3D reconstruction. The set of points in the global frame, obtained from the preceding method, is defined as $^Op_i = \{^Op_1, ^Op_2,\dots,^Op_n \}$, where $^Op_i \in R^3$. These points are integrated into the VDB volume as projective signed distances to each voxel. Each voxel at position $x$ is updated using a weighted cumulative signed distance, as described by the following equations. 
\begin{equation}
D_t(x)=\frac{W_{t-1}(x) D_{t-1}(x) + w_t(x) d_t(x)} {W_{t-1}(x)+w_t (x)}
\label{Eqn: distance grid}	
\end{equation}
\begin{equation}
W_t(x)= W_{t-1}(x) + w_t(x)
\label{Eqn: weight grid}
\end{equation}
where $D_{t-1}(x)$, $d_t(x)$, and $D_t(x)$ are previous, measured, and new signed distance values at voxel location $x$. Similarly, $W_{t-1}(x)$, $w_t(x)$, and $W_t(x)$ are previous, given and new weight values at voxel location x.


To locate the voxel to be updated, ray-casting is performed from the sensor origin to each newly acquired point. Subsequently, both the signed distance field and weight fields are updated. The precision of the reconstruction depends on the selected weight function.

Within the volumetric fusion framework, the weight function mitigates sensor noise and registration errors between sensors by smoothing distance values, as described in \autoref{Eqn: weight grid}. The cumulative weighted signed distance helps minimize noise from multiple range measurements. However, fusing range measurements from expanding regions tends to average the distance. To address this, an adaptive weighting strategy is implemented for dynamic scenes, ensuring an accurate representation of the build surface while minimizing sensor noise. The deposition location is tracked, and the area surrounding the deposition spot is designated as the active deposition region, while all other areas are considered inactive. The point cloud acquired from recent deposition in the active region is integrated into Voxels using \autoref{Eqn: weight active}, whereas points from the inactive region are integrated using \autoref{Eqn: weight inactive}. 

 

\begin{equation}
   w_A(d) = 
   \begin{cases}
        1, & \text{if $d<0$}\\
        0, & \text{if $d > \delta$}

   \end{cases}
   \label{Eqn: weight active}
\end{equation}

\begin{equation}
   w_I(d) = 
   \begin{cases}
        1, & \text{if $d<0$}\\
        1-d/\delta, & \text{if $0 < d < \delta$}\\
        0  & \text{if $d > \delta$}\\

   \end{cases}
    \label{Eqn: weight inactive}
\end{equation}	
where $w_A$ and $w_I$ are weight values as a function of signed distance ($d$) to be provided as a lambda expression.  
This approach efficiently fuses measurements from recently deposited areas while reducing noise impacts on stationary regions, enhancing its applicability in live additive manufacturing (AM). Consequently, the distance field representing the AM part surface is generated and updated with high confidence.
Finally, the implicit surface representation is converted into an explicit representation using a marching cube algorithm \citep{lorensen_marching_1987} by marching along all the cubes, adding a triangle to the list, and generating the final mesh as the union of all triangles.
\section{Deviation Mapping and  Monitoring}\label{sec: deviation}
\autoref{fig: defect detection} depicts the proposed method to generate a 3D deviation map and classify geometric defects on the fly. An innovative near-net shape model is formulated in \autoref{sec: reference mesh} that creates a reference model at any given time or layer, where its signed distance comparison with the 3D reconstructed surface of the actual part provides an accurate 3D deviation map. This 3D map is updated in real-time and then analyzed layer-by-layer to monitor defects, comprising detection and tracking of deviation growth and death as presented in \autoref{sec: geometric defect}. 

In near-net-shaped AM, a specified tolerance, denoted as $\delta_G$, defines the acceptable distance between the reference and actual measured geometry. We classify deviations exceeding $\delta_G$ as global deviations. Within this tolerance, certain local variations $\delta_L$ could also be observed in material deposition, which is classified under local over- or under-build regions. Although these local deviations are allowed individually, their cumulative effect over time can potentially result in a global deviation. Monitoring these local deviations provides essential early defect detection and process optimization insights. In the context of a complex deposition toolpath, it is essential to recognize that the deviation in the layer-by-layer deposition process can propagate and become more pronounced in subsequent layers \citep{chew_-process_2024,qin_geometric_2022}.


\vspace{-2mm}
\subsection{Novel near-net-shape reference model}\label{sec: reference mesh}
\begin{figure}[t]
    \centering
    \begin{subfigure}[b]{0.22\columnwidth}
         \centering 
         \includegraphics[width=\columnwidth]{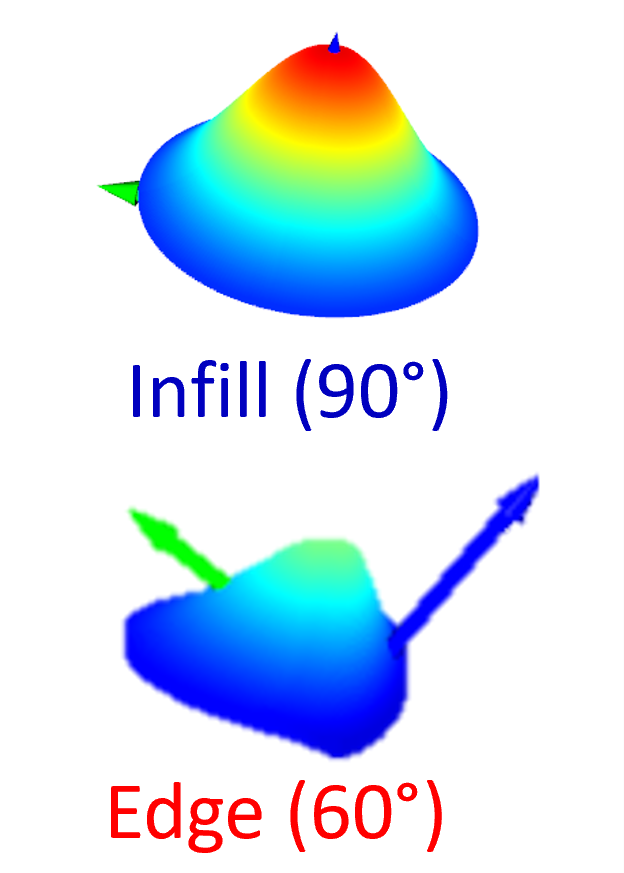}
         \caption{}
         \label{fig: gaussian}
    \end{subfigure}
    \hfill
    \begin{subfigure}[b]{0.34\columnwidth}
         \centering 
         \includegraphics[width=\columnwidth]{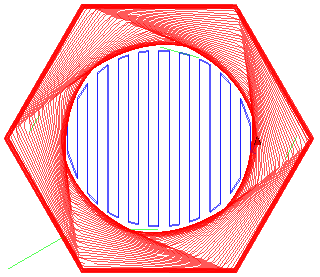}
         \caption{}
         \label{fig: toolpath_layer}
    \end{subfigure}
    \hfill
    \begin{subfigure}[b]{0.34\columnwidth}
         \centering 
         \includegraphics[width=\columnwidth]{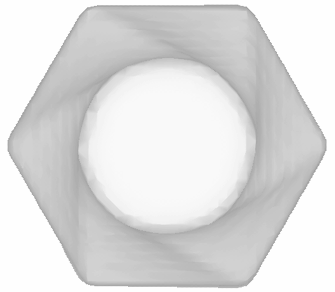}
         \caption{}
         \label{fig: reference_mesh}
    \end{subfigure}
    \caption{Near-net reference model generation: (a) A single spot deposition plume in perpendicular and off-normal angles. (b) A raster toolpath having infill and edge paths. Infill toolpath involves a perpendicular spray, and edge paths are off-angle spray (60$^{\circ}$). (c) A surface mesh is generated by combining Gaussian deposition spots up to a layer (i.e., layer 20 of the twisted tower). }
    \label{fig: gaussian reference model}
\end{figure}

We propose a novel reference model that accommodates the permissible geometric tolerances in the near-net shape instead of the strict CAD geometry. We adopt a Gaussian approximation and incrementally model the deposition distribution along the executed toolpath, \autoref{fig: toolpath_layer}. The reference model can be obtained at any time/layer \autoref{fig: reference_mesh}. 

As with other thermal spray technologies, out of nozzle jet distribution \citep{chen_modelling_2017}, CSAM coating thickness distribution \autoref{fig: gaussian} can be expressed by Gaussian approximation modelling as:
\vspace{-2mm}
\begin{equation}
   \phi = \zeta(\theta)\int_{0}^{T}{ \Biggl( \iint {\dfrac{A}{\sigma\sqrt{2\pi}} e^{-\left({\dfrac{x-\mu_x}{2\sigma^2}}+{\dfrac{y-\mu_y}{2\sigma^2}}\right)}}\,dx\,dy\Biggr)}\,dt
\end{equation}
Where A is the amplitude factor, $\sigma$ is the standard deviation, ($\mu$x, $\mu$y) is the center of the coating profile, and $\zeta(\theta)$ is the DE as a function of spray angle. Point clouds are integrated and fused into the volumetric grid by integrating the Gaussian deposition over time along the generated toolpath. The blue lines in \autoref{fig: toolpath_layer} are infill crosshatch patterns with perpendicular orientation, and the red lines are edge paths having a spray angle of 35$^{\circ}$. The green lines are skip moves where the robot speed is increased to 50mm/s. The cyan color represents an overhang structure where the traverse speed is decreased to 12 mm/s to deposit material outward. 
These toolpaths were generated using the Continuous3D software.
The same tool path a robot executes in the cold-spray process will be used in the mathematical model to create reference models. Intermediate near-net reference models are created for each layer, serving as the ground truth for comparison at every layer. These models facilitate not just layer-specific verification but also comparisons at any given time. For this paper, a layer-by-layer reference model has been systematically generated. As such, the model can also be extended using more sophisticated data-driven modelling approaches that can utilize the machine learning techniques \citep{ikeuchi_data-driven_2024}.

\subsection{Global and local under- and Over-built detection, segmentation and tracking}\label{sec: geometric defect}

The method depicted in \autoref{fig: defect detection} begins with creating the deviation map. Both reference and measured 3D models are converted to a 3D mesh using the same volumetric reconstruction method explained in \autoref{sec: dynamic}. Then, an online signed distance comparison between two surfaces at any layer or time provides the geometric deviation map between the desired and actual shape. For the signed distance comparison, we have represented both meshes in the same coordinate frame. The reference mesh is converted to an implicit surface representation. The closest distance of each vertex on the 3D-scanned mesh to the reference mesh surface is computed. The signed distance is negative or positive based on whether the query point is inside or outside the mesh.

Then the two-step defect detection and segmentation start with assessing global deviation. This initial phase involves a comparative analysis of the signed distance $d$  relative to a predefined global threshold $\delta_G$ across both the negative and the positive domains. Vertices exhibiting a signed distance exceeding $\delta_G$ are designated as overbuilds, whereas those with a signed distance inferior to $-\delta_G$ are labeled underbuilds. For the remaining vertices that fall within the confines of the global threshold, the procedure progresses to the subsequent phase, which encompasses the analysis of local deviations. The local threshold calculation is based on the mean curvature and signed distance principles. We have developed a new metric to evaluate local overbuild and underbuild conditions. The curvature value effectively represents both scenarios. By employing signed distance multiplication, we can differentiate between overbuild and underbuild situations.

It is essential to segment the mesh appropriately to analyze the size, shape, and features of the deviations. A distance threshold is employed to distinguish between the normal surface and the deviated surface. Based on the sign of the computed distances and utilizing the connectivity information inherent in the triangulated mesh, the connected edges are organized into groups. Following this, the segmented defects are monitored, and the surface area, along with the height of each deviation, is enclosed within a bounding box for further analysis.
Since the mesh is a connected component, it is then automatically analyzed and segmented, and the deviations are tracked, providing a defect change rate throughout the deposition process.



\section{Experiments and Results}\label{sec: experiment}
This section outlines the experiments conducted to validate in-process 3D reconstruction, deviation segmentation, and tracking techniques. In practical AM applications, it is common to fabricate complex geometries rather than simple shapes, including single-track, multi-track, or thin-wall structures. Two intricate models were selected to examine in-process reconstruction and defect detection within cold spray AM: twisted towers and seven-sided shapes.

\begin{figure}[h]
    \centering
    \begin{subfigure}[b]{\columnwidth}
        \centering
        \includegraphics[width=.9\textwidth]{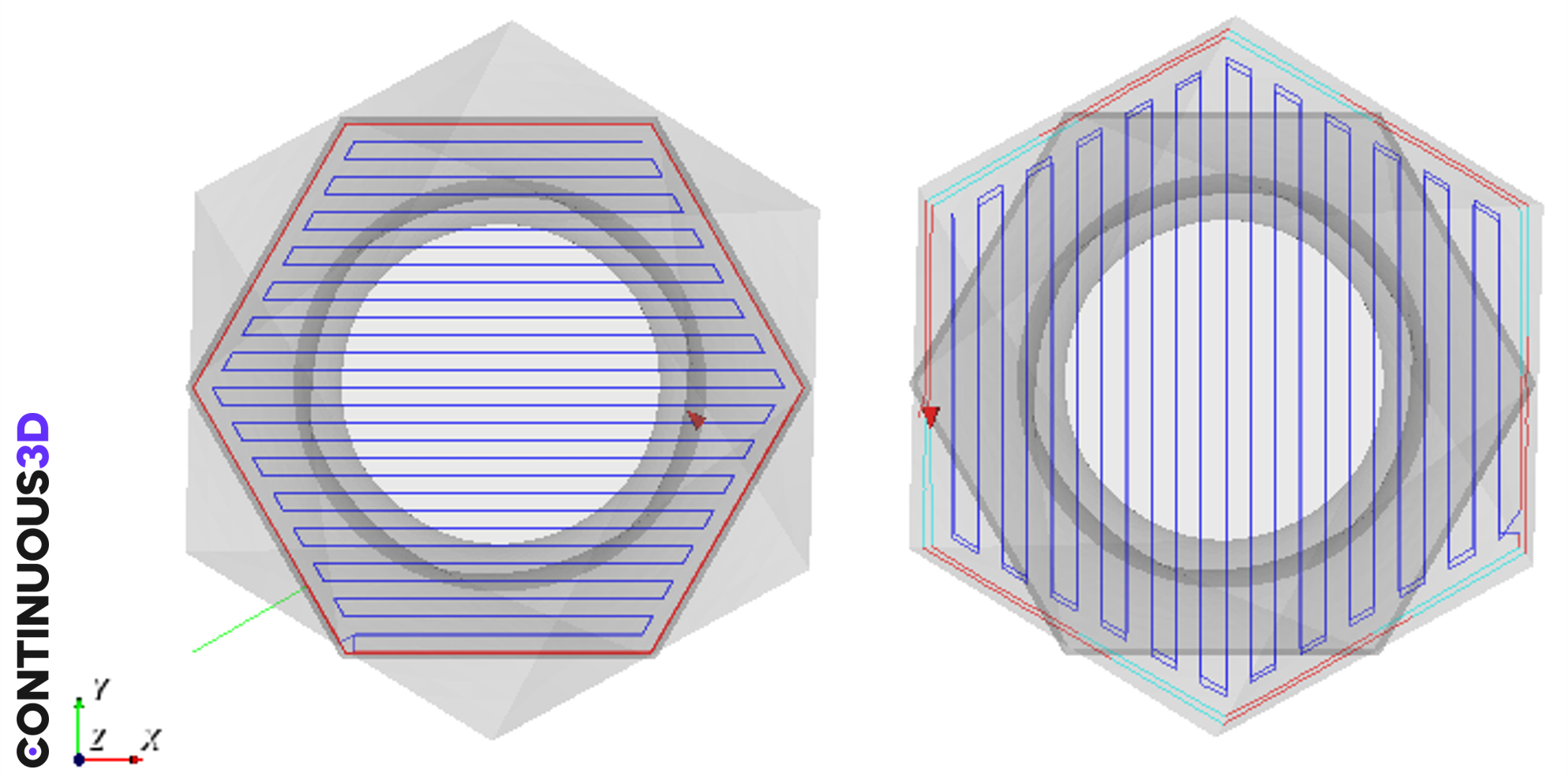}
        \caption{ Toolpath of layers 1 and 62 (left to right).}
        \label{fig: TT_toolpath}
    \end{subfigure}
    \begin{subfigure}[b]{\columnwidth}
        \centering
        \includegraphics[width=.9\textwidth]{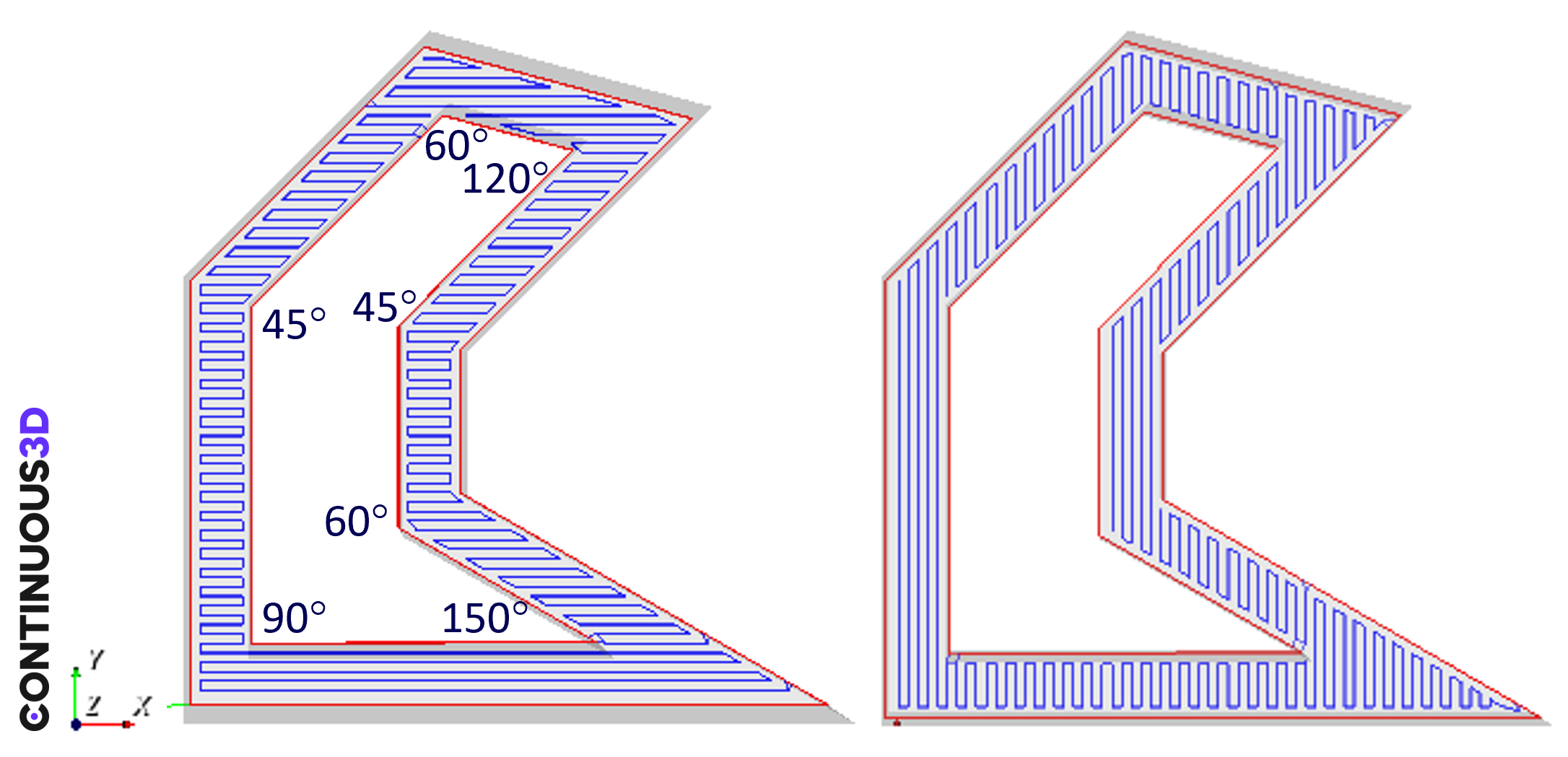}
        \caption{ Toolpath of layers 1 and 12 (left to right).}
        \label{fig: SS_toolpath}
    \end{subfigure}
    \caption{A continuous toolpath generated using Continuous3D software Version 1.0.95 for a twisted tower and seven-sided shape. The toolpath type is color-coded (\textcolor{red}{edge}, \textcolor{blue}{infill}, \textcolor{green}{skip}, and \textcolor{cyan}{overbuild}).}
    \vspace{-0.5cm}
    \label{fig: toolpath}
\end{figure}


\def \figwidth{0.49}
\begin{figure}[h]
    \centering
    \begin{subfigure}[b]{\figwidth\columnwidth}
        \centering
        \includegraphics[width=\textwidth, trim={0 4mm 0 7mm},clip]{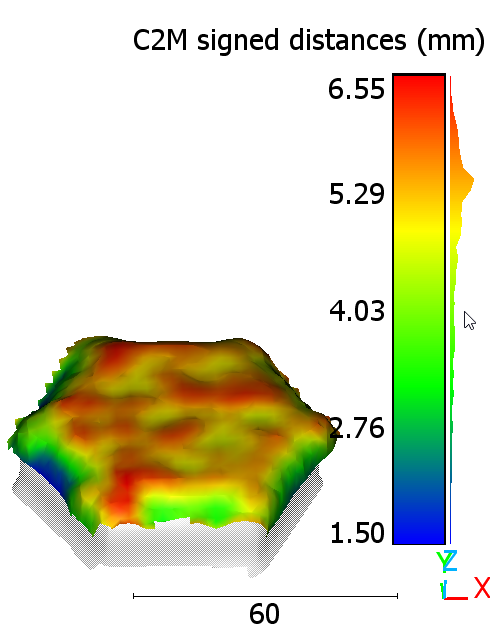}
        \caption{Layer 10}
        \label{fig: Exp1_layer10_c2m}
    \end{subfigure}
    \begin{subfigure}[b]{\figwidth\columnwidth}
        \centering
        \includegraphics[width=\textwidth, trim={0 4mm 0 7mm},clip]{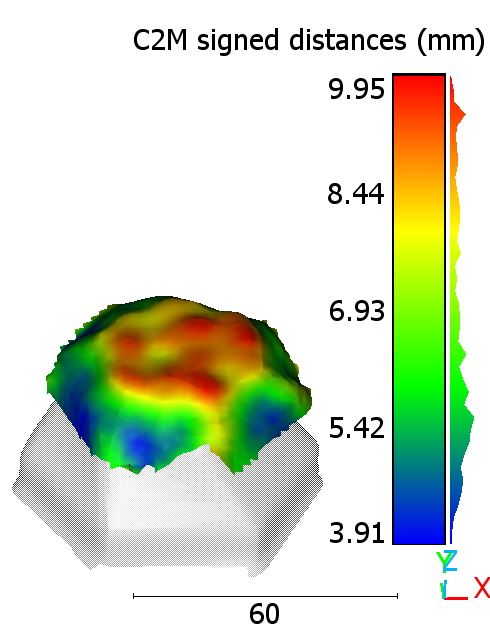}
        \caption{Layer 20}
        \label{fig: Exp1_layer20_c2m}
    \end{subfigure}
    \begin{subfigure}[b]{\figwidth\columnwidth}
        \centering
        \includegraphics[width=\textwidth, trim={0 4mm 0 7mm},clip]{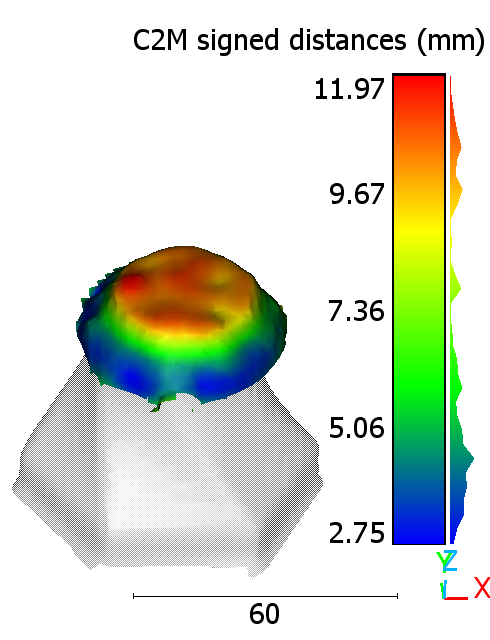}
        \caption{Layer 30}
        \label{fig: Exp1_layer30_c2m}
    \end{subfigure}
    \begin{subfigure}[b]{\figwidth\columnwidth}
        \centering
        \includegraphics[width=\textwidth, trim={0 4mm 0 7mm},clip]{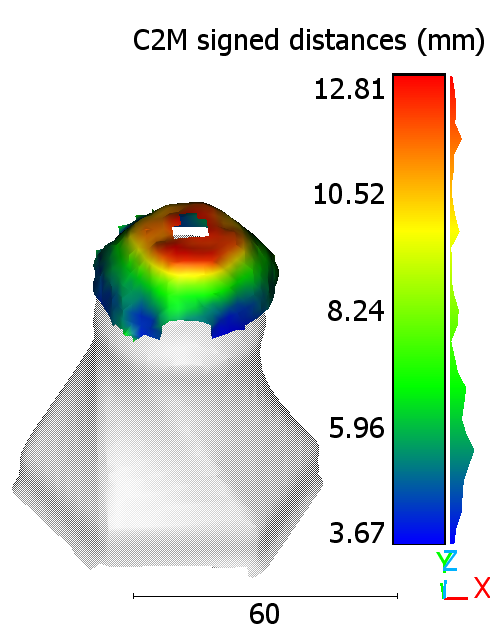}
        \caption{Layer 40}
        \label{fig: Exp1_layer40_c2m}
    \end{subfigure}
    \begin{subfigure}[b]{\figwidth\columnwidth}
        \centering
        \includegraphics[width=\textwidth, trim={0 4mm 0 7mm},clip]{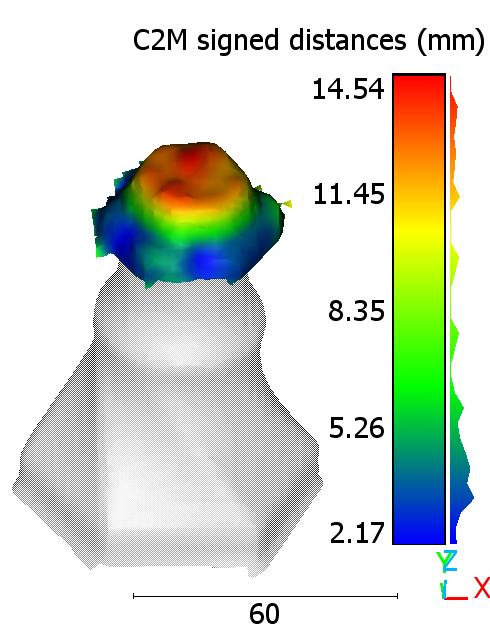}
        \caption{Layer 50}
        \label{fig: Exp1_layer50_c2m}
    \end{subfigure}
    \begin{subfigure}[b]{\figwidth\columnwidth}
        \centering
        \includegraphics[width=\textwidth, trim={0 4mm 0 7mm},clip]{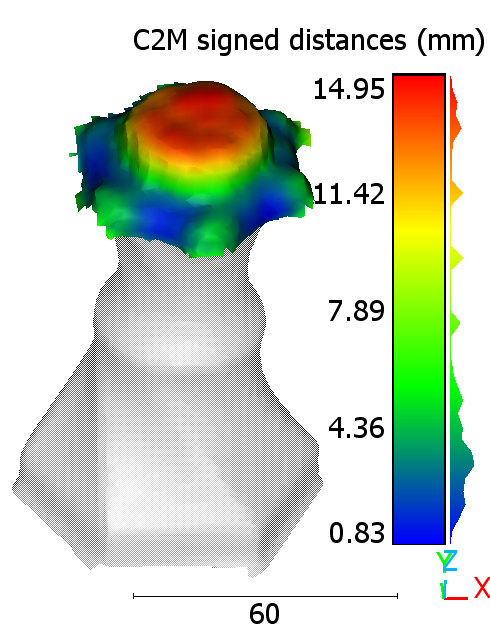}
        \caption{Layer 60}
        \label{fig: Exp1_layer60_c2m}
    \end{subfigure}
    \caption{Experiment 1: Layerwise analysis of twisted tower dataset. Global deviation/Part height deviation was observed in (a) Layer 10, (b) Layer 20, (c) Layer 30, (d) Layer 40, (e) Layer 50, and (f) Layer 60.}
    \label{fig: Exp1}
\end{figure}
 \begin{figure}[h]
    \centering
    \begin{subfigure}[b]{\figwidth\columnwidth}
        \centering
        \includegraphics[width=\textwidth, trim={0 4mm 0 7mm},clip]{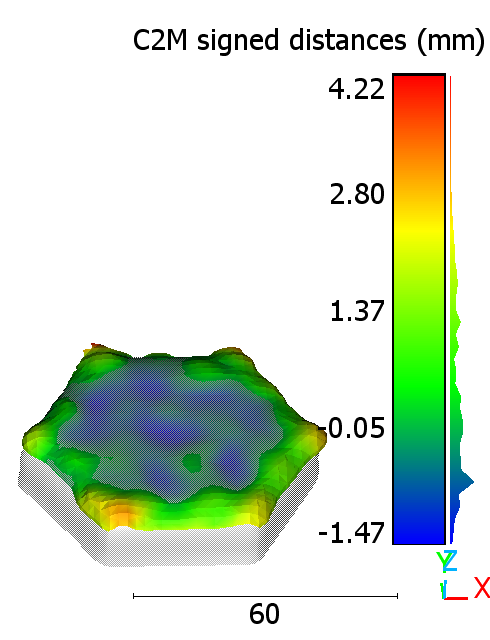}
        \caption{Layer 10}
        \label{fig: Exp2_layer10_c2m}
    \end{subfigure}
    \begin{subfigure}[b]{\figwidth\columnwidth}
        \centering
        \includegraphics[width=\textwidth, trim={0 4mm 0 7mm},clip]{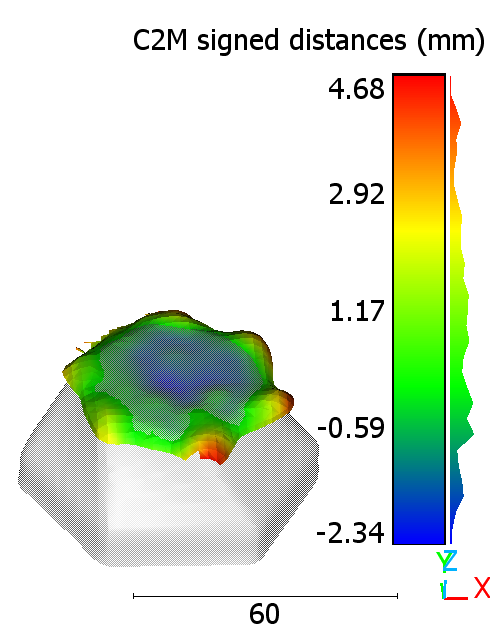}
        \caption{Layer 20}
        \label{fig: Exp2_layer20_c2m}
    \end{subfigure}
    \begin{subfigure}[b]{\figwidth\columnwidth}
        \centering
        \includegraphics[width=\textwidth, trim={0 4mm 0 7mm},clip]{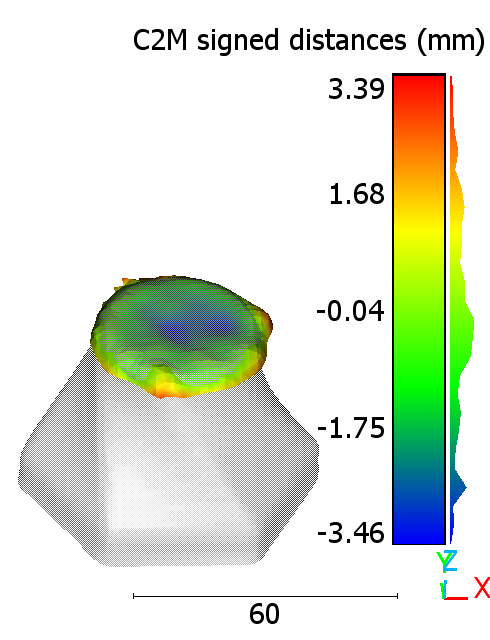}
        \caption{Layer 30}
        \label{fig: Exp2_layer30_c2m}
    \end{subfigure}
    \begin{subfigure}[b]{\figwidth\columnwidth}
        \centering
        \includegraphics[width=\textwidth, trim={0 4mm 0 7mm},clip]{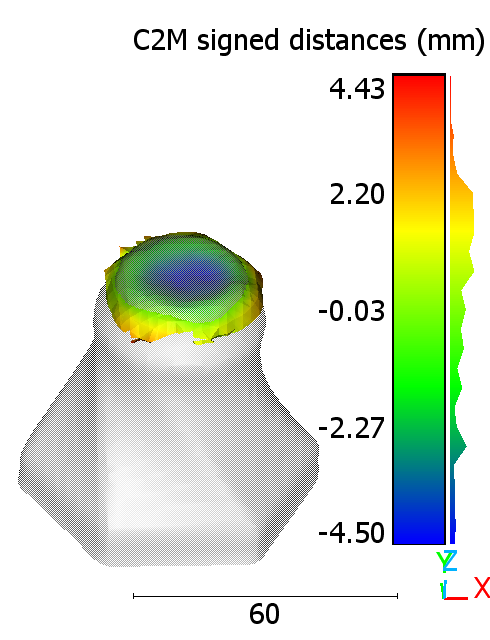}
        \caption{Layer 40}
        \label{fig: Exp2_layer40_c2m}
    \end{subfigure}
    \begin{subfigure}[b]{\figwidth\columnwidth}
        \centering
        \includegraphics[width=\textwidth, trim={0 4mm 0 7mm},clip]{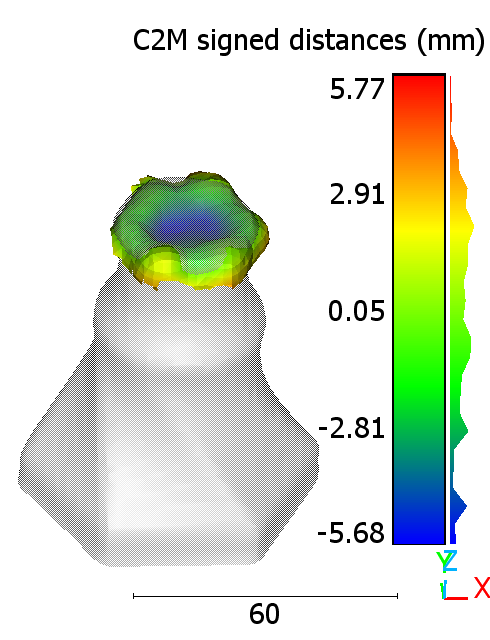}
        \caption{Layer 50}
        \label{fig: Exp2_layer50_c2m}
    \end{subfigure}
    \begin{subfigure}[b]{\figwidth\columnwidth}
        \centering
        \includegraphics[width=\textwidth, trim={0 4mm 0 7mm},clip]{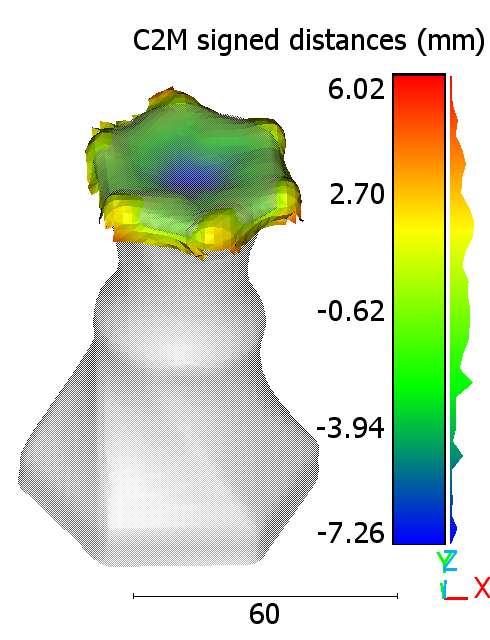}
        \caption{Layer 60}
        \label{fig: Exp2_layer60_c2m}
    \end{subfigure}
    \caption{Experiment 2: Layerwise analysis of twisted tower dataset. Global deviation/Part height deviation was observed in (a) Layer 10, (b) Layer 20, (c) Layer 30, (d) Layer 40, (e) Layer 50, and (f) Layer 60. }
    \label{fig: Exp2}
\end{figure}

\def \figwidthExpThree{0.49}
\begin{figure}[t]
    \centering
    \begin{subfigure}[b]{\figwidthExpThree\columnwidth}
        \centering
        \includegraphics[width=\textwidth,trim={0 4mm 0 7mm},clip]{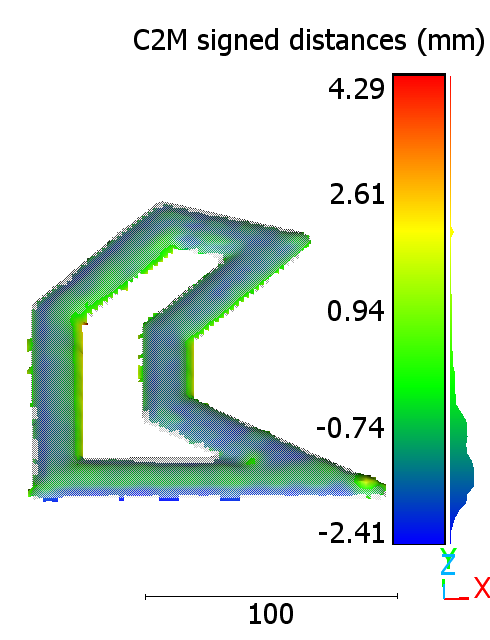}
        \caption{Layer 3}
        \label{fig: Exp3_layer3_c2m}
    \end{subfigure}
    \begin{subfigure}[b]{\figwidthExpThree\columnwidth}
        \centering
        \includegraphics[width=\textwidth,trim={0 4mm 0 7mm},clip]{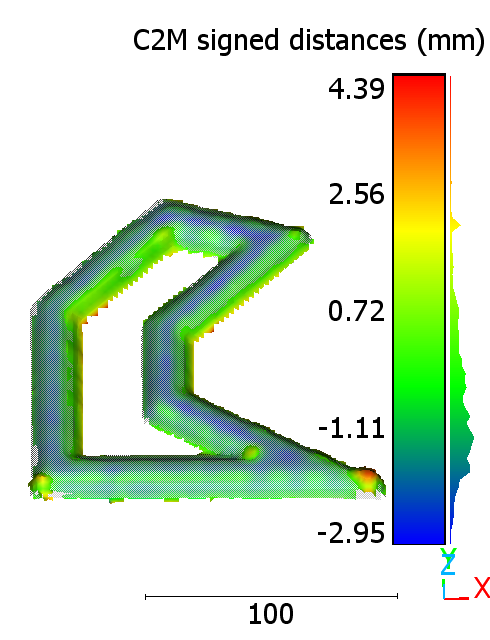}
        \caption{Layer 6}
        \label{fig: Exp3_layer6_c2m}
    \end{subfigure}
    \begin{subfigure}[b]{\figwidthExpThree\columnwidth}
        \centering
        \includegraphics[width=\textwidth,trim={0 4mm 0 7mm},clip]{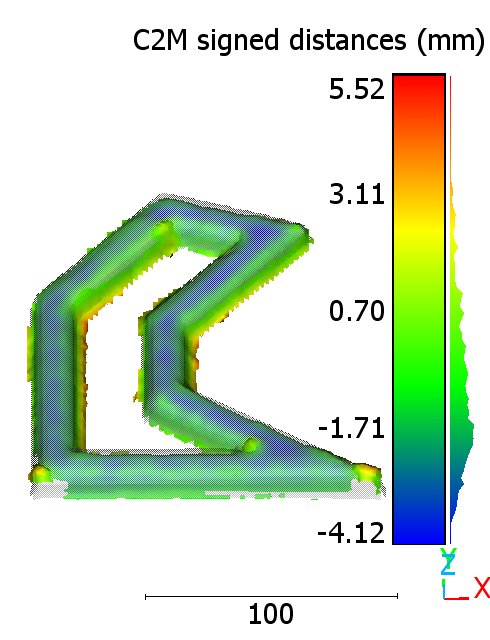}
        \caption{Layer 9}
        \label{fig: Exp3_layer9_c2m}
    \end{subfigure}
    \caption{Experiment 3: Layerwise analysis of a seven-sided dataset based on signed distance comparison with near-net reference model.}
    \label{fig: Exp3}
\end{figure}
\def \figwidthExpFour{0.49}
\begin{figure}[ht]
    \centering
    \begin{subfigure}[b]{\figwidthExpFour\columnwidth}
        \centering
        \includegraphics[width=\textwidth,trim={0 4mm 0 7mm},clip]{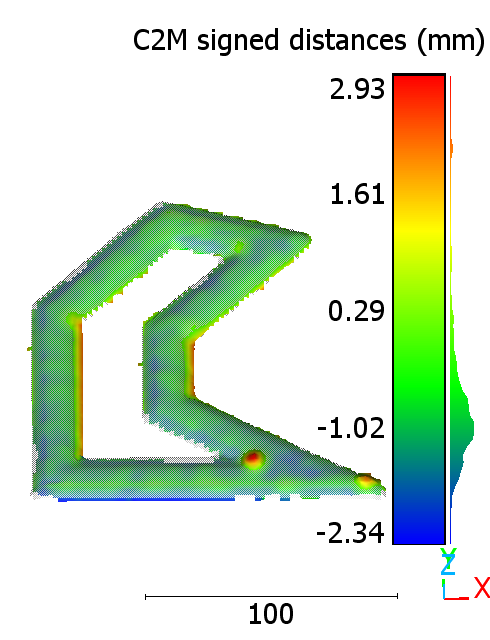}
        \caption{Layer 3}
        \label{fig: Exp4_layer3_c2m}
    \end{subfigure}
    \begin{subfigure}[b]{\figwidthExpFour\columnwidth}
        \centering
        \includegraphics[width=\textwidth,trim={0 4mm 0 7mm},clip]{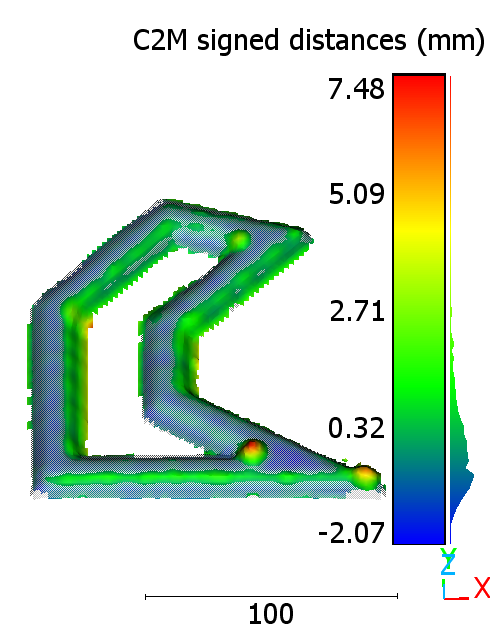}
        \caption{Layer 6}
        \label{fig:Exp4_layer6_c2m}
    \end{subfigure}
    \begin{subfigure}[b]{\figwidthExpFour\columnwidth}
        \centering
        \includegraphics[width=\textwidth,trim={0 4mm 0 7mm},clip]{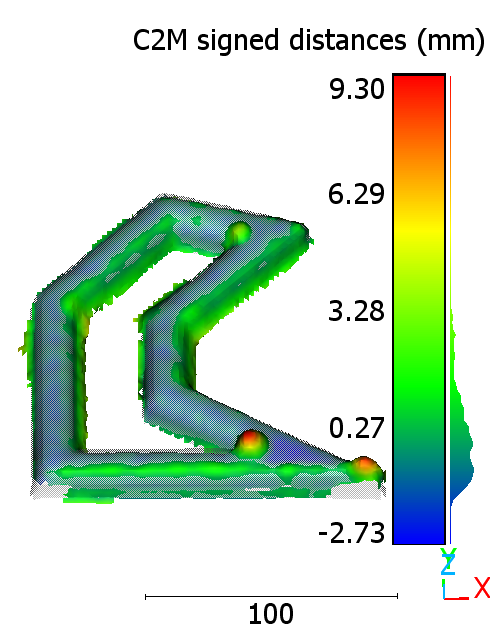}
        \caption{Layer 9}
        \label{fig: Exp4_layer9_c2m}
    \end{subfigure}
    \begin{subfigure}[b]{\figwidthExpFour\columnwidth}
        \centering
        \includegraphics[width=\textwidth,trim={0 4mm 0 7mm},clip]{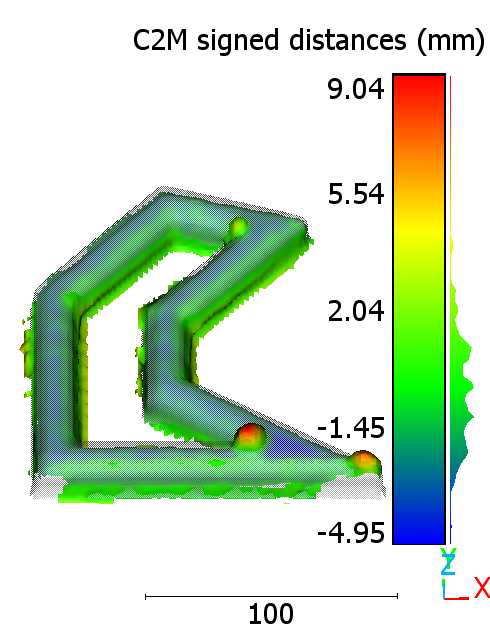}
        \caption{Layer 12}
        \label{fig: Exp4_layer12_c2m}
    \end{subfigure}
    \caption{Experiment 4: Layerwise analysis of a seven-sided dataset based on signed distance comparison with near-net reference model.}
    \label{fig: Exp4}
\end{figure}

The experiments employed the PCS 1000L model cold spray gun manufactured by Plasma Giken Co. Ltd. in Japan. The system includes a 268mm long water-cooled tungsten carbide nozzle featuring a 15 mm inlet diameter, 3mm throat diameter, and 6.55mm exit diameter. A commercial gas-atomised titanium grade 2 powder (CP-Ti) with a size distribution ranging from 15 to 45 $\mu$m was used as a deposition material, procured from AP\&C in Boisbriand, Canada.

To initiate the system, nitrogen gas was passed through the nozzle to establish and sustain a pressure of 5 MPa while maintaining a temperature of \num{900}~\si{\degreeCelsius} and a gas flow rate of \num{2050}~SLM (Standard Litres per Minute). The carrier gas flow rate was maintained at 300 SLM. The stirrer of the feeder was disabled to enable the free flow of powder into the system. The stand-off distance (SOD), defined as the distance from the tip of the nozzle to the point of deposition, was established at an initial value of 30 mm.

The toolpath generated is shown in \autoref{fig: toolpath}. The toolpath parameters were selected based on the expert knowledge. The line spacing was 2 mm, and the layer thickness was calculated to be 0.8 mm. The traverse speed of the deposition was different based on the type of toolpath: infill-30mm/s, contour-30mm/s, overhang-12mm/s, and skip-50mm/s. The contour angle was set to $\ang{35}$, and a zone radius of 1mm was used. Once the system was stable, the powder feeder was turned on with a rotational speed of 0.5 rpm, and the robot program waited one minute before starting to maintain the powder flow. The point cloud processing, 3D reconstruction, and deviation identification method were implemented using software written in Python. Open3d \citep{zhou_open3d_2018}  and VDBFusion \citep{vizzo_vdbfusion_2022} libraries were used for point cloud processing and 3D reconstruction. The pipeline was implemented on an Intel Xeon W-2133 Processor with 3.6 GHz and 64 GB RAM, installed with an Nvidia Quadro P2200 GPU with 5 GB of memory.


\subsection{Layerwise analysis}
\subsubsection{Twisted tower}
The twisted tower represents a complex geometric structure distinguished by its multi-track and multi-layer design. This configuration embodies a practical and fully realized commercial component, unlike a single-track experimental segment often used in experiments.  The shape features a non-uniform cross-section along the printing direction, with sections that twist and narrow toward the center. From there, the sections expand to form an overhanging structure. This case study employed a continuous double-layer strategy with cross-hatched toolpath planning. In this double-layer approach, two infill paths are utilized—one moving forward and the other in reverse—followed by two contour paths. During the contour path, the cold spray nozzle is tilted to minimize tapering and create straight walls.


It is important to note that two experiments were conducted using the same toolpath and parameters to print this shape. However, these experiments resulted in completely different builds. Experiment 1 resulted in a global overbuild as shown in \autoref{fig: Exp1}, illustrated as the signed distance from the reference model. Whereas, Experiment 2 exhibited some surface roughness at start and transitioned to a global underbuild starting from layer 20, as shown in \autoref{fig: Exp2}. This shows the limitation of the current open-loop system, lacking repeatability and reliability of the HDRRAM process, and emphasizing the need for the proposed system to detect these deviations early, so that future research could focus on predictive or reactive control systems.
  
\begin{table}[t]
    \centering
    \caption{Twisted tower experiments final height and deposited weight.}
    \label{tab: table1}
    \begin{tabular}{p{0.1\columnwidth}p{0.1\columnwidth}p{0.1\columnwidth}p{0.1\columnwidth}p{0.1\columnwidth}p{0.15\columnwidth}} 
    \toprule
         Exp.no&  Total edge height (mm)&  Average edge height (mm)&  Total center height (mm)&  Average center height (mm)& Deposition weight (g)\\ 
    \midrule
         Exp 1&  101.13&  0.82&  114.87&  0.93& 819.23\\ 
         Exp 2&  99.60&  0.80&  90.60&  0.73& 687.31\\
    \bottomrule
    \end{tabular}\vspace{-0.3cm}
\end{table}
\begin{table}[t]
    \caption{Seven-sided shape experiments final height and deposited weight.}
    \label{tab: table2}%
    \begin{tabular}{p{0.15\columnwidth}p{0.2\columnwidth}p{0.1\columnwidth}p{0.1\columnwidth}p{0.15\columnwidth}}
    \toprule
        Exp. no & Strategy &  Total edge height (mm)  & Avg edge height (mm) & Deposition weight (g) \\
    \midrule
        Exp3 & Double layer & 11.00 & 0.73 & 510.87\\
        Exp4 & Single layer & 17.20 & 0.82 & 424.00\\
    \bottomrule
    \end{tabular}
\end{table}
\subsubsection{Seven-sided shape}
A seven-sided geometric shape has been developed to investigate the impact of toolpath angles on deposited geometry and the occurrence of geometric defects. The design incorporates various angles, specifically 45$^{\circ}$, 60$^{\circ}$, 90$^{\circ}$, 120$^{\circ}$, 150$^{\circ}$, along the sides of the robot path, as shown in \autoref{fig: toolpath}. The shape is 24 mm in height and 13 mm in width, featuring a hollow center.


The operation is sustained without interruption in the context of a continuous deposition process, specifically the CSAM method. However, the point cloud data is methodically divided into distinct layers to optimize memory utilization. Fluctuations in the height of the tool govern the transitions between these layers. The layerwise analysis focused on identifying defects as they develop at the end of each layer of part fabrication. A signed distance comparison between two meshes provides a deviation result. For experiment 3 with the double-layer strategy, A signed distance from the 3D scanned model to the reference model is shown in \autoref{fig: Exp3}. \autoref{fig: Exp4} shows layerwise comparison for experiment 4, at layers 3, 6, 9, and 12.

\subsection{Automated geometric deviation detection and segmentation pipeline}\label{sec: pipeline}


\begin{figure}[h]
    \centering
    \begin{subfigure}[b]{\figwidth\columnwidth}
        \centering
        \includegraphics[width=\columnwidth, trim={2cm 0 2cm 5cm},clip]{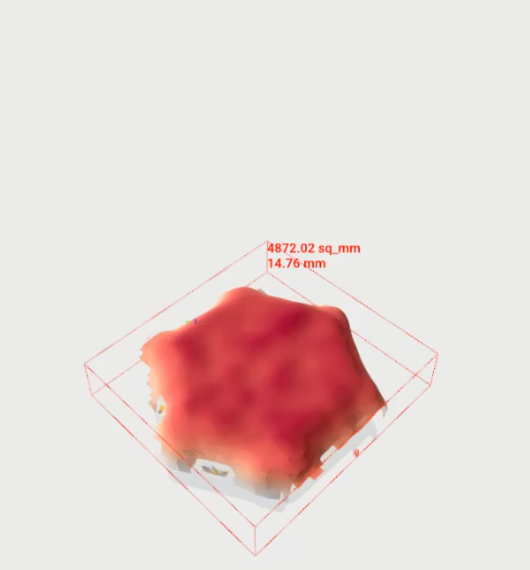}
        \caption{Layer 10}
        \label{fig: Exp1_seg_track_10}
    \end{subfigure}
    \begin{subfigure}[b]{\figwidth\columnwidth}
        \centering
        \includegraphics[width=\columnwidth,trim={2cm 0 2cm 5cm},clip]{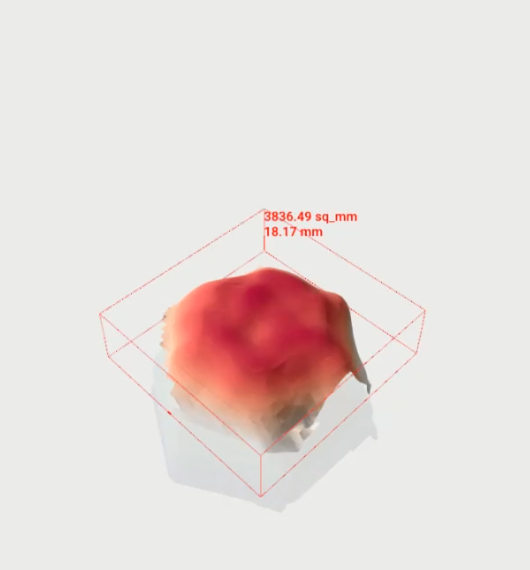}
        \caption{Layer 20}
        \label{fig: Exp1_seg_track_20}
    \end{subfigure}
    \begin{subfigure}[b]{\figwidth\columnwidth}
        \centering
        \includegraphics[width=\columnwidth,trim={2cm 1cm 2cm 35mm},clip]{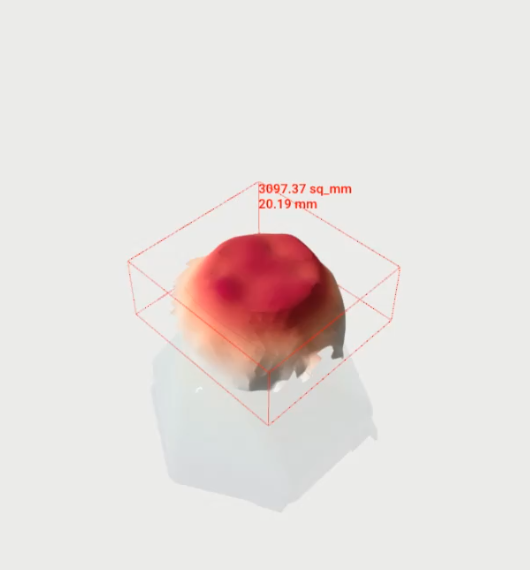}
        \caption{Layer 30}
        \label{fig: Exp1_seg_track_30}
    \end{subfigure}
    \begin{subfigure}[b]{\figwidth\columnwidth}
        \centering
        \includegraphics[width=\columnwidth,trim={2cm 1cm 2cm 35mm},clip]{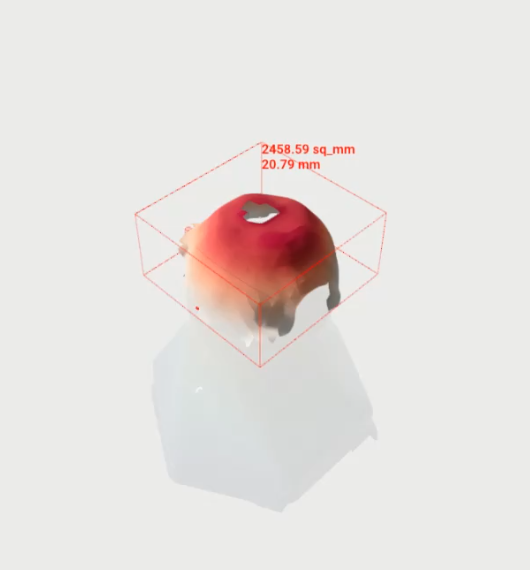}
       \caption{Layer 40}
        \label{fig: Exp1_seg_track_40}
    \end{subfigure}
    \begin{subfigure}[b]{\figwidth\columnwidth}
        \centering
        \includegraphics[width=\columnwidth,trim={2cm 1cm 2cm 0},clip]{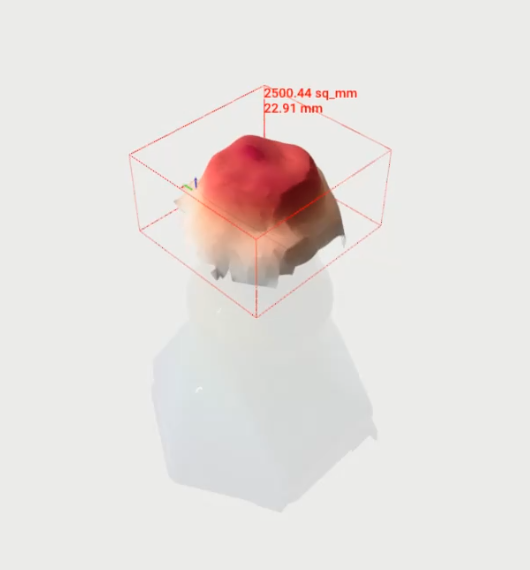}
        \caption{Layer 50}
        \label{fig: Exp1_seg_track_50}
    \end{subfigure}
    \begin{subfigure}[b]{\figwidth\columnwidth}
        \centering
        \includegraphics[width=\columnwidth,trim={2cm 1cm 2cm 0},clip]{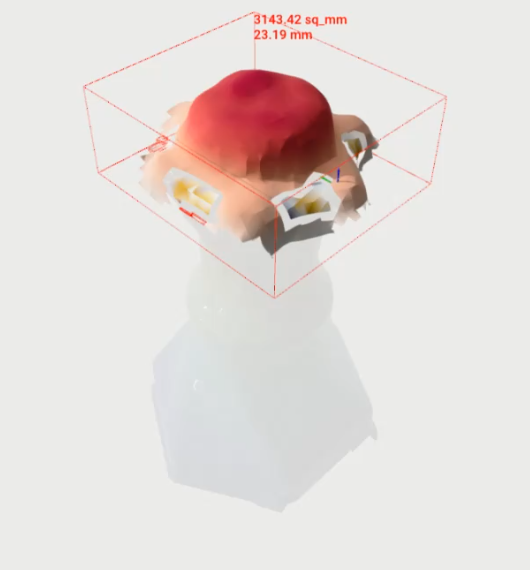}
       \caption{Layer 60}
        \label{fig: Exp1_seg_track_60}
    \end{subfigure}
    \caption{Experiment 1: Automated defect segmentation and tracking (a) Global deviation/Part height deviation was observed in (a) Layer 10, (b) Layer 20, (c) Layer 30, (d) Layer 40, (e) Layer 50, and (f) Layer 60.}
    \label{fig: Exp1_seg_track}
\end{figure}
\begin{figure}[h]
    \centering
    \includegraphics[width=\linewidth]{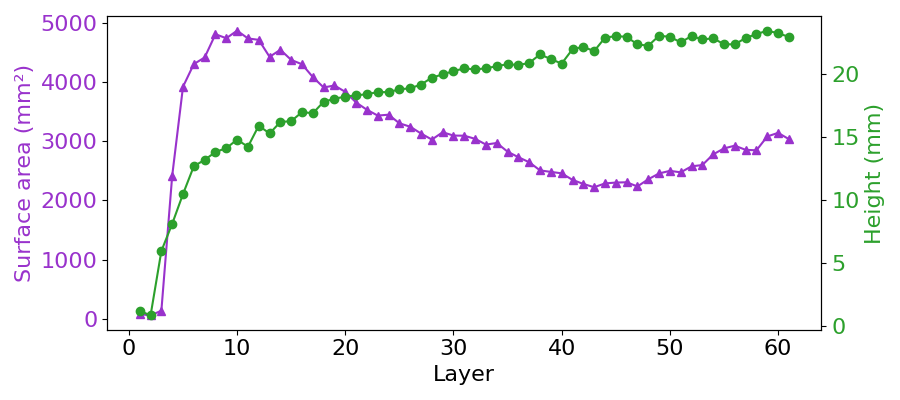}
    \caption{Evolution of overbuild defect in terms of surface area and defect in experiment 1.}
    \label{fig: Exp1_area_height}
\end{figure}

\begin{figure}[h]
    \centering
    \begin{subfigure}[b]{\figwidth\columnwidth}
        \centering
        \includegraphics[width=\columnwidth,trim={2cm 1cm 2cm 6cm},clip]{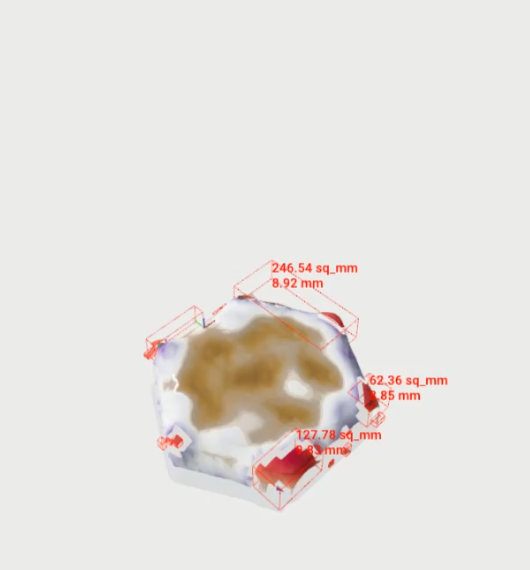}
        \caption{Layer 10}
        \label{fig: Exp2_seg_track_10}
    \end{subfigure}
    \begin{subfigure}[b]{\figwidth\columnwidth}
        \centering
        \includegraphics[width=\columnwidth,trim={2cm 1cm 2cm 6cm},clip]{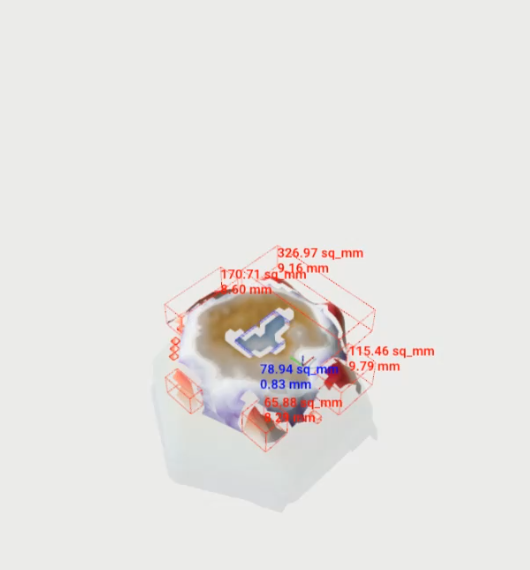}
        \caption{Layer 20}
        \label{fig: Exp2_seg_track_20}
    \end{subfigure}
    \begin{subfigure}[b]{\figwidth\columnwidth}
        \centering
        \includegraphics[width=\columnwidth,trim={2cm 1cm 2cm 5cm},clip]{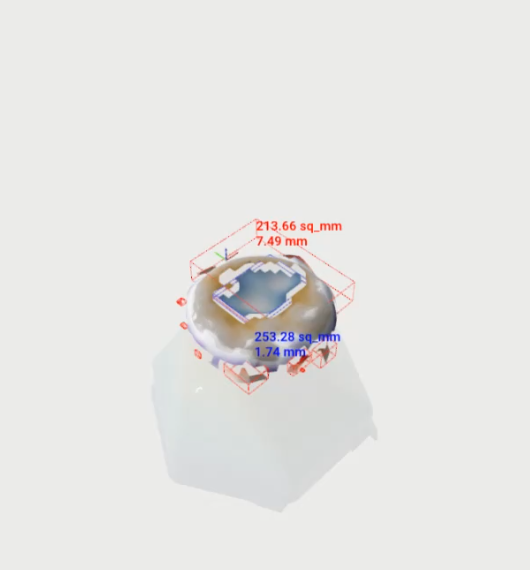}
        \caption{Layer 30}
        \label{fig: Exp2_seg_track_30}
    \end{subfigure}
    \begin{subfigure}[b]{\figwidth\columnwidth}
        \centering
        \includegraphics[width=\columnwidth,trim={2cm 1cm 2cm 5cm},clip]{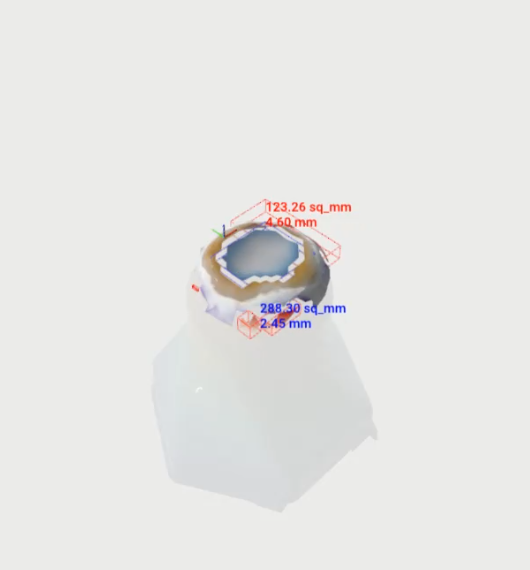}
        \caption{Layer 40}
        \label{fig: Exp2_seg_track_40}
    \end{subfigure}
    \begin{subfigure}[b]{\figwidth\columnwidth}
        \centering
        \includegraphics[width=\columnwidth,trim={2cm 1cm 2cm 15mm},clip]{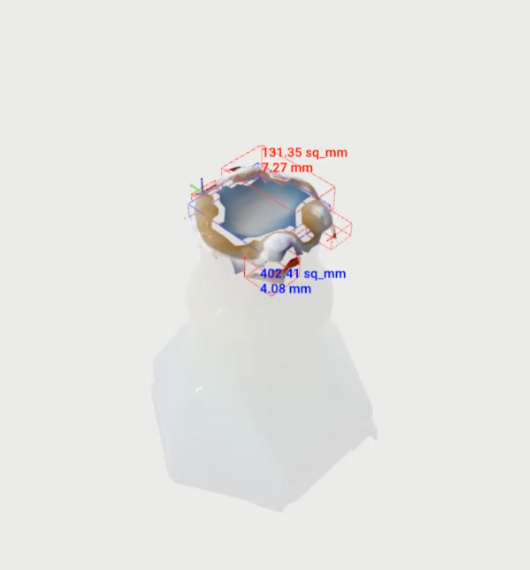}
        \caption{Layer 50}
        \label{fig: Exp2_seg_track_50}
    \end{subfigure}
    \begin{subfigure}[b]{\figwidth\columnwidth}
        \centering
        \includegraphics[width=\columnwidth,trim={2cm 1cm 2cm 15mm},clip]{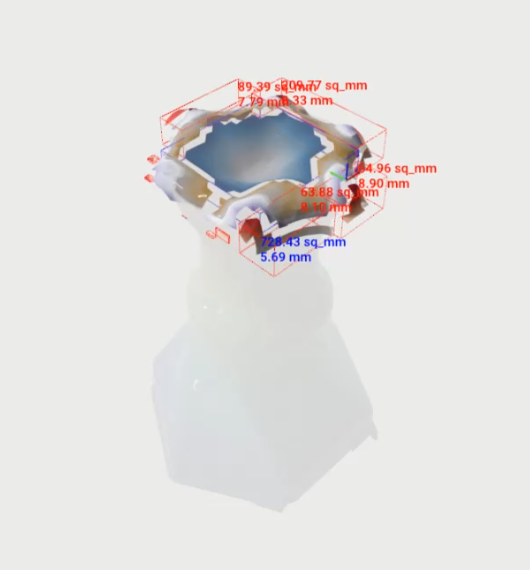}
        \caption{Layer 60}
        \label{fig: Exp2_seg_track_60}
    \end{subfigure}
    \caption{Experiment 2: Automated defect segmentation and tracking (a) Global deviation/Part height deviation and local deviation were observed in (a)  Layer 10, (b)  Layer 20, (c) Layer 30, (d) Layer 40, (e) Layer 50, and (f) Layer 60.}
    \label{fig: Exp2_seg_track}
\end{figure}


\begin{figure}[h]
    \centering
    \includegraphics[width=\columnwidth]{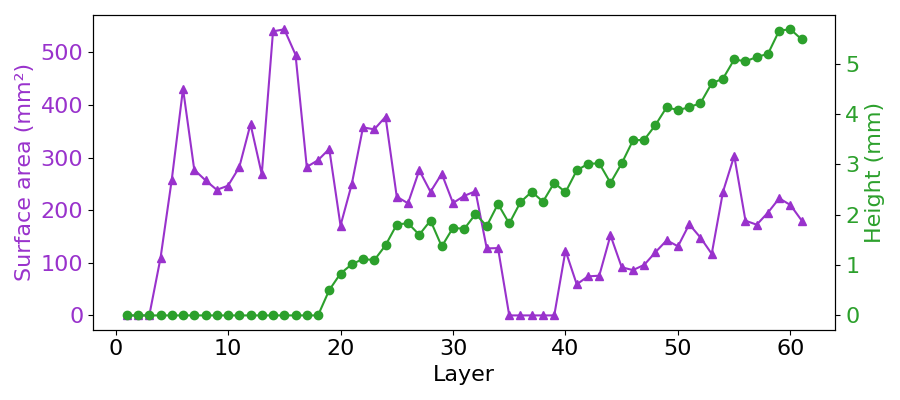}
    \caption{Evolution of underbuild defect in terms of surface area and defect in experiment 2. }
    \label{fig: Exp2_area_height_tc}
\end{figure}

The automated geometric defect segmentation and tracking pipeline in Cold Spray Additive Manufacturing (CSAM) comprises several critical components. It initiates with a multi-laser scanning system that captures a three-dimensional point cloud in real-time. The acquired data points are systematically filtered, transformed, and integrated into a volumetric grid. Subsequently, a surface mesh is extracted and compared against a near-net reference shape for each layer. Geometric defects are identified by applying a signed-distance metric, and tracking these defects facilitates the observation of their evolution across successive layers. This robust pipeline effectively identifies geometric defects, segments regions of overbuild and underbuild, and enables the continuous assessment of process integrity within the CSAM framework.
\hypersetup{hyperfootnotes=false}
\renewcommand{\thefootnote}{\alph{footnote}}
\renewcommand{\footnoterule}{\vspace{1mm}\hrule height 0.5pt width 5cm \vspace{2mm}}

The geometric deviations are detected as explained in \autoref{sec: deviation}. The vertices are flagged as overbuild, underbuild, or normal. (Here, overbuild and underbuild are referred to as global and normal vertices within the global threshold $\delta_G$.) The new mesh is created from the selected vertices and color-coded\footnotemark \space based on the signed distance value. Furthermore, for the vertices within the global threshold limit, a new mesh is formed, and the signed distance is calculated. A new local surface deviation is calculated using a curvature-weighted signed distance metric and color-coded on the normalized metric value.
The global deviations are tracked based on the intersection of the bounding box. 

\footnotetext{The global overbuild and underbuild are shown by sequential red and blue colors, respectively. The local surface deviation is shown with an orange-purple sequential color.}

\subsubsection{Twisted tower}
The automated deviation segmentation and tracking pipeline has successfully identified a global overbuild, which has been monitored across the various layers. \autoref{fig: Exp1_seg_track} illustrates the progression of the overbuild for layers 10, 20, 30, 40, 50, and 60. The deviated surface mesh is contained within the axis-aligned bounding box, with annotations indicating both the surface area of the mesh and the height of the bounding box. Additionally, \autoref{fig: Exp1_area_height} provides a detailed summary of the surface area and height for each layer. It is observed that the surface area is directly proportional to the part's cross-section, validating the global threshold detected by the algorithm. At the same time, the height of the overbuild exhibits an upward trend with the addition of each successive layer, illustrating the propagation of defects layer-by-layer.

Likewise, \autoref{fig: Exp2_seg_track} presents the progression of underbuild across layers 10, 20, 30, 40, 50, and 60. \autoref{fig: Exp2_area_height_tc} complements this by detailing the surface area and height for each respective layer. The analysis indicates a negative global deviation of the twisted tower layers, as detected by the algorithm. Notably, the surface area is reduced in the middle layers, while it increases in the lower layers. This pattern continues from the middle to the top, reflecting the cross-sectional geometry of the part.

\subsubsection{Seven-sided shape}
\begin{figure}[h]
    \centering
    \includegraphics[width=0.99\columnwidth]{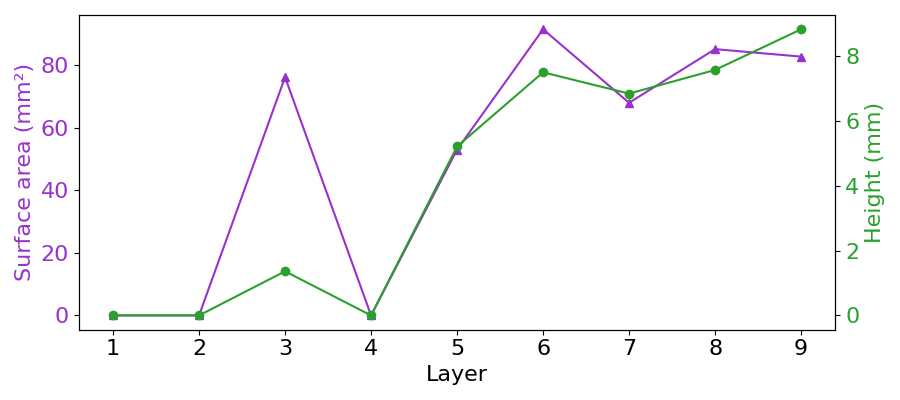}
    \caption{A significant localised deviation around the corner with $\ang{150}$ of the seven-sided shape in experiment 3 in terms of surface area and height.}
    \label{fig: Exp3_area_height_tc}
\end{figure}

\begin{figure}[h]
    \centering
    \begin{subfigure}[b]{0.49\textwidth}
        \centering
        \includegraphics[width=\textwidth]{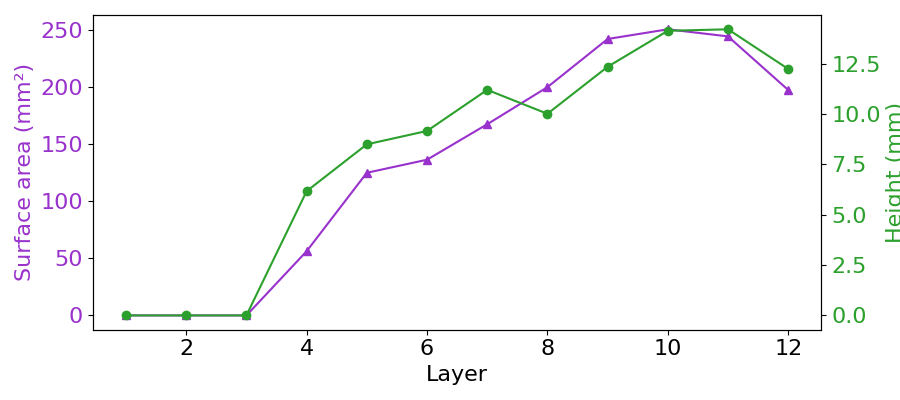}
        \caption{}
        \label{fig: Exp4_area_height_C3_tc}
    \end{subfigure}
    \begin{subfigure}[b]{0.49\textwidth}
        \centering
        \includegraphics[width=\textwidth]{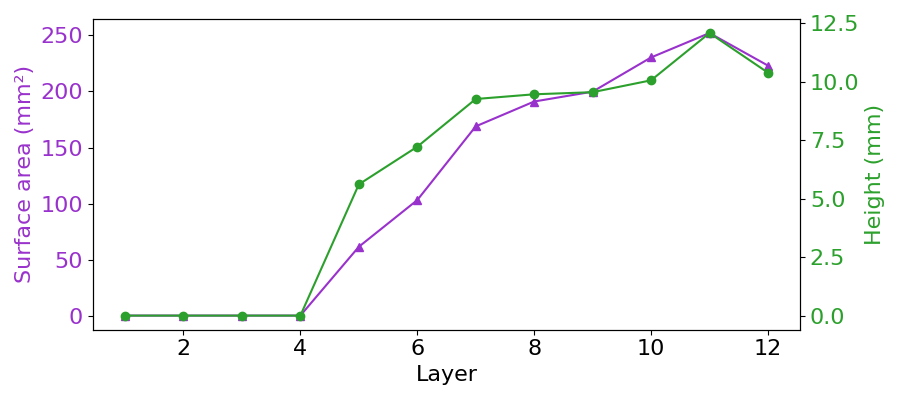}
        \caption{}
        \label{fig: Exp4_area_height_C4_tc}
    \end{subfigure}
    \caption{A significant positive deviation in the local area of the seven-sided at both inside and outside corners with $150^o$ angle.}
    \label{fig: Exp4_area_height}
\end{figure}

The automated deviation segmentation and tracking pipeline detected a global overbuild that is tracked throughout the layers. \autoref{fig: Exp3_seg_track} presents the evolution of overbuild for layers 3, 6, and 9. The deviated surface mesh is within the axis-aligned bounding box, with the text displaying the surface area of the mesh and the height of the bounding box. \autoref{fig: Exp3_area_height_tc} presents the surface area and height for each layer. The analysis indicates a significant localised deviation around the corner with $\ang{150}$ of the seven-sided shape. It is observed that both the surface area and height of the underbuild exhibit an upward trend with the addition of each successive layer.
Similarly, \autoref{fig: Exp4_seg_track} presents the evolution of overbuild for layers 3, 6, 9, and 12. The surface area and height of the deviation text are displayed, and the segmented defect is contained within a bounding box, as shown in \autoref{fig: Exp4_area_height}. The analysis indicates a significant positive deviation in the local area of the twisted tower layers. It is observed that both the surface area and height of the underbuild exhibit an upward trend with the addition of each successive layer. (a) indicates the overbuild of the outer bottom right corner, and (b) represents an overbuild of the inner bottom right corner (see \autoref{fig: Exp4_seg_track})

\def \figwidthExpThreeSeg{0.49}
\begin{figure}[h]
    \centering
    \begin{subfigure}[b]{\figwidthExpThreeSeg\columnwidth}
        \centering
        \includegraphics[width=\columnwidth]{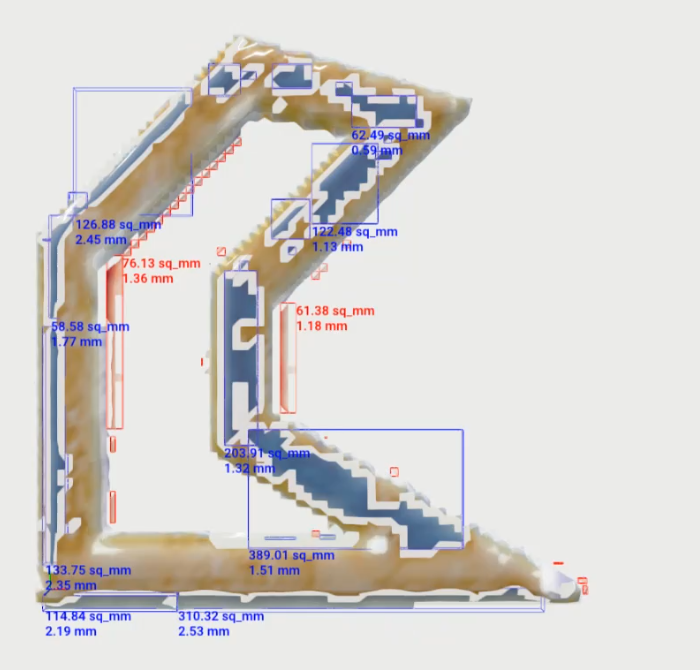}
        \caption{Layer 3}
        \label{fig: Exp3_seg_track_3}
    \end{subfigure}
    \begin{subfigure}[b]{\figwidthExpThreeSeg\columnwidth}
        \centering
        \includegraphics[width=\columnwidth]{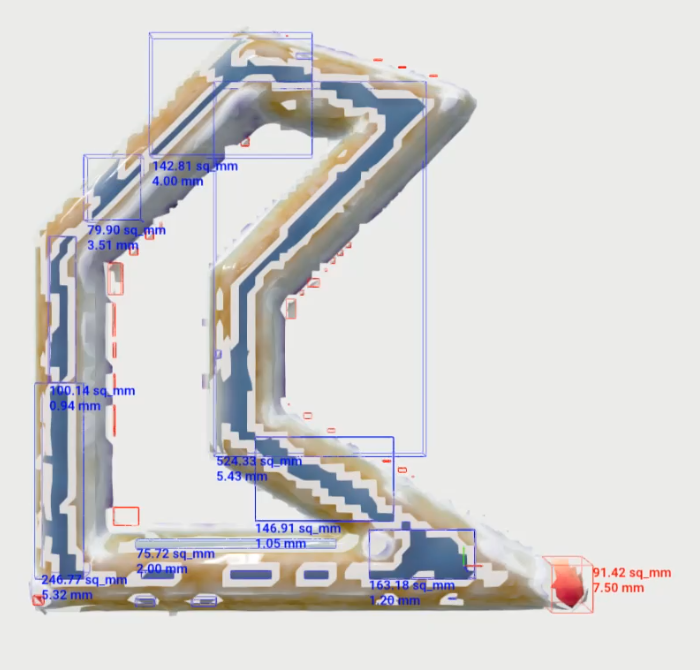}
        \caption{Layer 6}
        \label{fig: Exp3_seg_track_6}
    \end{subfigure}
    \begin{subfigure}[b]{\figwidthExpThreeSeg\columnwidth}
        \centering
        \includegraphics[width=\columnwidth]{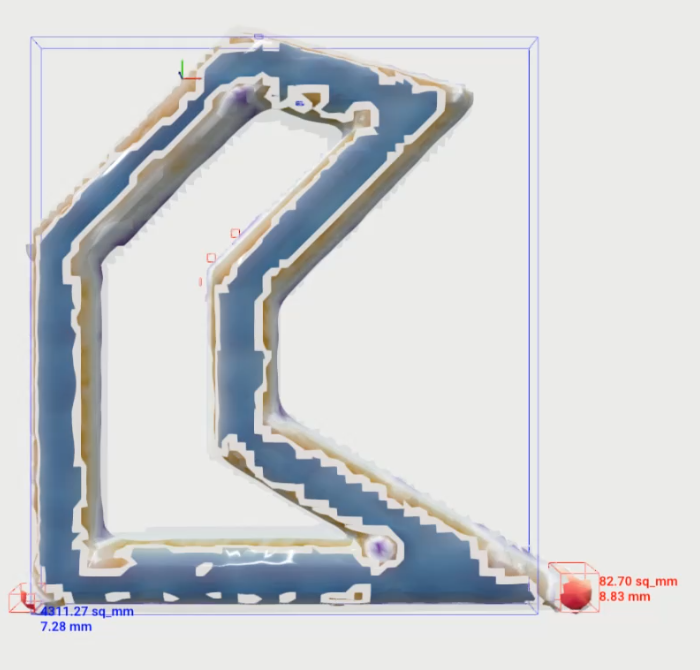}
        \caption{Layer 9}
        \label{fig: Exp3_seg_track_9}
    \end{subfigure}
    \caption{Experiment 3: Automated defect segmentation and tracking. (a) Global deviation/Part height deviation and local deviation were observed in layers 3, 6, and 9. }
    \label{fig: Exp3_seg_track}
\end{figure}

\def \figWidthExpFourSeg{0.49}
\begin{figure}[h]
    \centering
    \begin{subfigure}[b]{\figWidthExpFourSeg\columnwidth}
        \centering
        \includegraphics[width=\columnwidth]{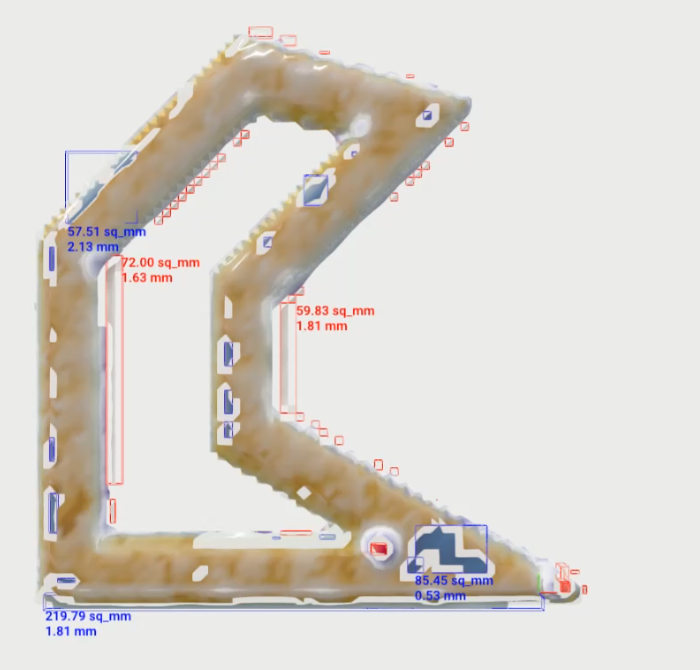}
        \caption{Layer 3}
        \label{fig: Exp4_seg_track_3}
    \end{subfigure}
    \begin{subfigure}[b]{\figWidthExpFourSeg\columnwidth}
        \centering
        \includegraphics[width=\columnwidth]{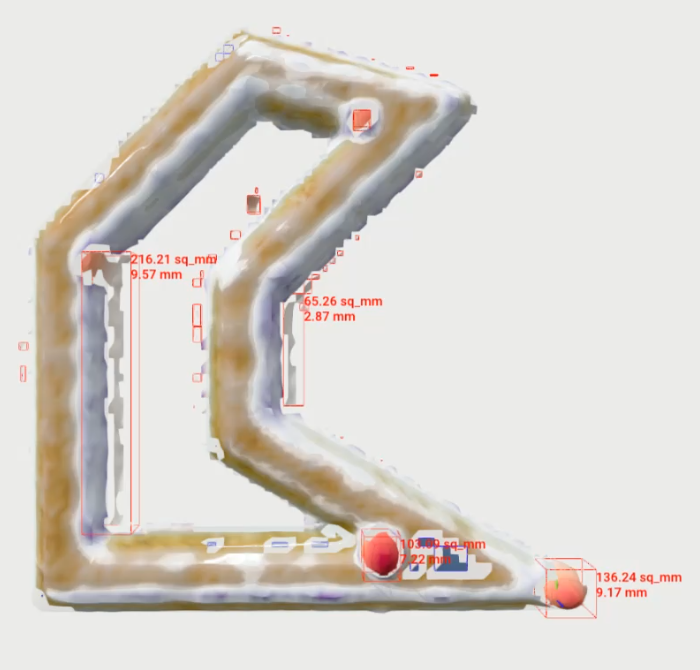}
        \caption{Layer 6}
        \label{fig: Exp4_seg_track_6}
    \end{subfigure}
    \begin{subfigure}[b]{\figWidthExpFourSeg\columnwidth}
        \centering
        \includegraphics[width=\columnwidth]{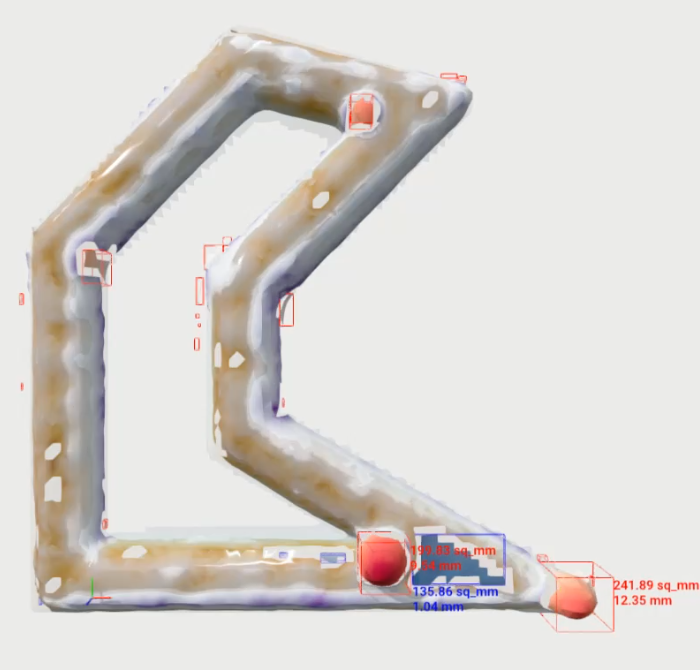}
        \caption{Layer 9}
        \label{fig: Exp4_seg_track_9}
    \end{subfigure}
    \begin{subfigure}[b]{\figWidthExpFourSeg\columnwidth}
        \centering
        \includegraphics[width=\columnwidth]{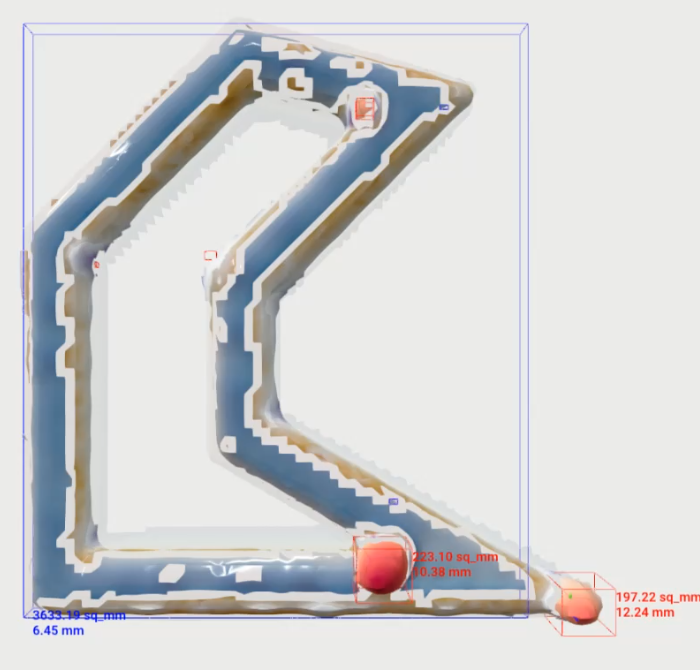}
        \caption{Layer 12}
        \label{fig: Exp4_seg_track_12}
    \end{subfigure}
    \caption{Experiment 4: Automated defect segmentation and tracking. (a) Global deviation/Part height deviation and local deviation were observed in layers 3, 6, 9, and 12. }
    \label{fig: Exp4_seg_track}
\end{figure}


\subsection{Validation with post-scanned 3D model as ground truth}
To ensure the robustness and accuracy of the proposed methodology, validation is conducted using a post-scanned 3D model as the ground truth. The ground truth model is acquired through a high-resolution  MetraScan3D scanner that performs metrology-grade measurements and 3D geometrical surface inspections with 0.025 mm resolution. The ground truth was captured after completing the deposition and letting the substrate and part cool down. 
The final layer of the reconstructed surface is compared against the ground truth to get a deviation map, as shown in \autoref{fig: Validation}. Ideally, the deviation needs to be minimal for higher reconstruction accuracy. However, the deviation in the result is due to the following reasons.
 \begin{itemize}
    \item The sparse, raw data due to hardware implementation limitations.
    \item A voxel size of 2 mm is used to reconstruct the surface
    \item The post-scanned mesh (stippled mesh) is not watertight, and there are missing surfaces.
    \item warping of the substrate due to residual stress.
 \end{itemize}

\def \figWidthVal {0.49}
\begin{figure}[h]
    \centering
    \begin{subfigure}[b]{\figWidthVal\columnwidth}
        \centering
        \includegraphics[width=\columnwidth]{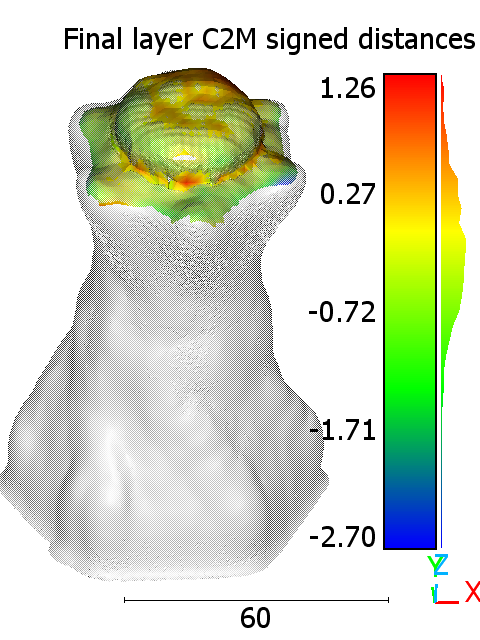}
        \caption{Experiment 1}
        \label{fig: Exp1_validation}
    \end{subfigure}
    \begin{subfigure}[b]{\figWidthVal\columnwidth}
        \centering
        \includegraphics[width=\columnwidth]{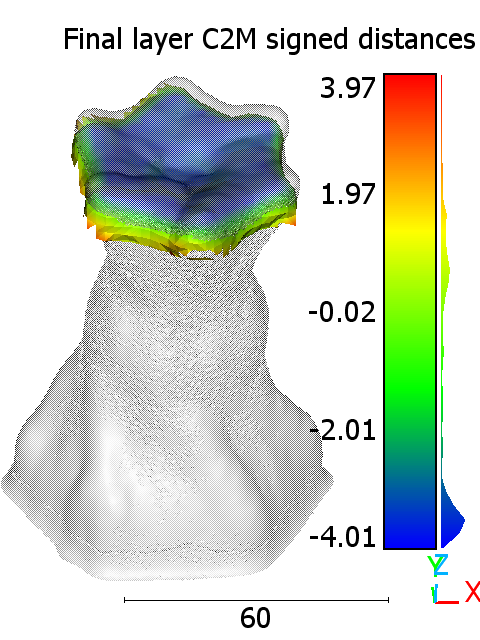}
        \caption{Experiment 2}
        \label{fig: Exp2_validation}
    \end{subfigure}
    \begin{subfigure}[b]{\figWidthVal\columnwidth}
        \centering
        \includegraphics[width=\columnwidth]{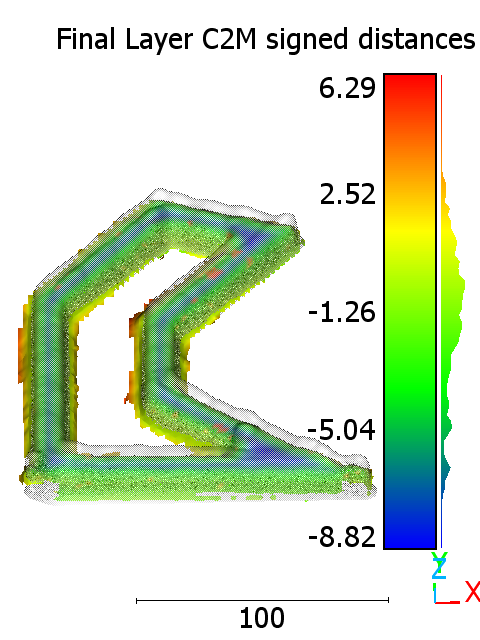}
        \caption{Experiment 3}
        \label{fig: Exp3_validation}
    \end{subfigure}
    \begin{subfigure}[b]{\figWidthVal\columnwidth}
        \centering
        \includegraphics[width=\columnwidth]{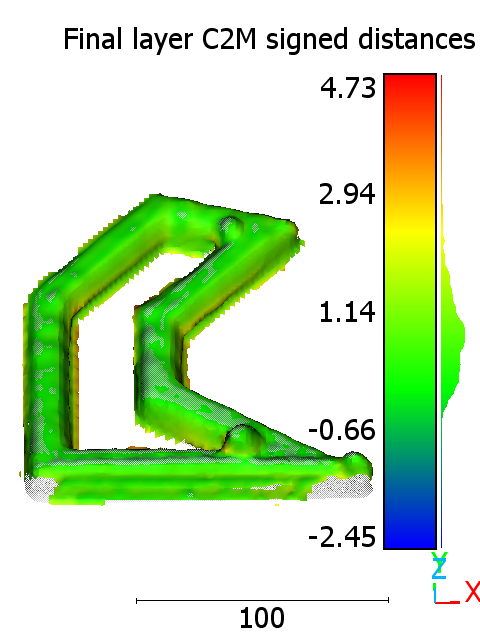}
        \caption{Experiment 4}
        \label{fig: Exp4_validation}
    \end{subfigure}
    \caption{Validation of reconstructed surface with post-scanned mesh. A twisted tower experiment (a) Experiment no. 1 (b) Experiment no. 2 and Seven-sided shape experiment (c)Experiment no.3 (A large deviation is there because the data acquisition process halted prematurely) and (d) Experiment no. 4}
    \label{fig: Validation}
\end{figure}

\section{Discussion}\label{sec: discussion}
The main result of this study is the development of a dynamic surface reconstruction framework offering a detailed virtual reconstruction of the physical build in real-time that is capable of mapping in-process deviations in dimensional accuracy on a layer-by-layer basis with high resolution. Beyond accurate monitoring, the 3D-DM$^2$ has the potential to be expanded for triggering corrective actions such as halting the build, pausing material feed, or modifying process parameters, minimizing the risk of cumulative errors and material waste.

The multi-scanner vision system was demonstrated to capture dynamic part growth in real-time to detect and track the local and global geometric defects in the CSAM process with the proposed in-process 3D deviation mapping and defect monitoring method. The method currently compares the in-process surface mesh to the reference mesh layer-by-layer. Comparing two surfaces using ray casting is a computationally expensive operation, and performing it layerwise doesn't interrupt the process; however, extracting the mesh and comparing it in real-time is challenging. The method can be scaled up to work in real-time by comparing two voxel grids instead of an extracted surface for a straightforward and efficient comparison, providing real-time capability for the pipeline in the future. 

In cold spray additive manufacturing (CSAM), once defects such as overbuild and underbuild are identified, this information can be used to develop a control strategy that compensates for these irregularities and improves deposition accuracy. CSAM, which relies on the high-velocity impact of solid particles rather than melting, benefits significantly from a closed-loop control system that integrates real-time monitoring, data analysis, and dynamic process adjustments. Parameters such as nozzle traverse speed, standoff distance, particle velocity, and powder feed rate can be modified in response to detected deviations. For example, reducing the powder feed rate or increasing traverse speed in regions with overbuild, or slowing down the nozzle or increasing the feed in areas with underbuild, helps maintain consistent layer geometry. Furthermore, machine learning techniques can be applied to predict defect-prone regions based on prior builds and optimize process parameters proactively.

Looking forward, a key area for future research lies in integrating real-time corrective feedback directly into the deposition process. Such integration would enable the system to go beyond discrete event responses and instead adaptively optimize parameters on-the-fly, improving process stability and part quality in real-time. The architecture developed in this study provides a strong foundation for such advancements and is designed to be generalisable across various high-throughput robotic additive manufacturing systems. 

\section{Conclusion}\label{sec: conclusion}
This paper presents the 3D deviation mapping and defect monitoring method in the CSAM process using a multi-scanner 3D scanning system. The dynamic surface construction subsystem reconstructs the complete 3D model of part growth. A 3D deviation map is obtained in real-time by utilizing a near-net reference shape model.  The method automatically analyzes the deviation map to segment geometric defects and track the local and global geometric defects. An experimental study showed the effectiveness of the proposed method to automatically detect and track the local and global deviation in comparison to the near-net reference shape.
This method is essential for detecting defects, providing a critical first step toward understanding their root causes. By expanding the approach through detailed analysis and data collection, it's possible to identify the underlying issues driving these defects. Building on this foundation, the system can then incorporate machine learning and adaptive control mechanisms to enable predictive insights and dynamic process adjustments. This progression not only helps prevent defects before they occur but also enhances the overall efficiency and quality of the manufacturing process.


\backmatter

\bmhead{Supplementary information}

\bmhead{Acknowledgments}
The authors want to acknowledge Mr. Robert Dark for his software implementation of a robotic scanner for experimental testing and validation, and Mr. Dang Quan Nguyen from Data61 for providing 3D scanning support for validation. Subash Gautam thanks the Royal Melbourne Institute of Technology (RMIT) and the Commonwealth Scientific and Industrial Research Organization (CSIRO) for his PhD scholarship through the CSIRO Active Integrated Matter - Future Science Platform (AIM-FSP). CSIRO’s Future Digital Manufacturing Fund (FDMF) support is also kindly acknowledged. 

\section*{Declarations}
\begin{itemize}
\item Funding:  Active Integrated Matter-Future Science Platforms (AIM-FSP), Future Digital Manufacturing Fund (FDMF)
\item Conflict of interest: None
\item Ethics approval and consent to participate: Not applicable
\item Consent for publication: Not applicable
\item Data availability: Not applicable
\item Materials availability: Not applicable
\item Code availability: Not applicable
\item     Author contribution\\
    \textbf{Subash Gautam}: Conceptualization, Methodology, Mathematical formulation, Data curation, Software,  Validation, Visualization, Writing-Original draft
    \textbf{Alejandro Vargas-Uscategui}: Conceptualization, Methodology, Investigation, Resources, Supervision, Writing-review and editing.
    \textbf{Peter King}: Conceptualization, Writing-review and editing
    \textbf{Alireza Bab-Hadiashar}: Supervision, Writing-review and editing
    \textbf{Ivan Cole}: Supervision, Writing-review, and editing. 
    \textbf{Ehsan Asadi}: Conceptualization, Methodology, Validation, Supervision, Writing-review and editing
    
\end{itemize}
\bibliography{sn-bibliography-z}

\end{document}